\documentclass[acmsmall,acmcsur]{acmart}
\usepackage{multirow}
\usepackage{color}
\usepackage[caption=false,font=footnotesize]{subfig}
\usepackage{array}
\usepackage{graphicx}
\usepackage{textcomp}
\usepackage{lipsum}
\usepackage[figuresright]{rotating}
\usepackage{adjustbox}
\usepackage{bm}

\usepackage{pifont}
\usepackage{ragged2e} 
\usepackage{makecell}

\AtBeginDocument{%
  \providecommand\BibTeX{{%
    \normalfont B\kern-0.5em{\scshape i\kern-0.25em b}\kern-0.8em\TeX}}}

\setcopyright{acmcopyright}
\copyrightyear{2025}
\acmYear{2025}
\acmDOI{10.1145/1122445.1122456}

\acmConference[CSUR]{ACM Computing Surveys}
\acmBooktitle{ACM Computing Surveys}
\acmPrice{15.00}
\acmISBN{978-1-4503-XXXX-X/18/06}

\begin{document}

\title [A Comprehensive Survey on Machine Learning Driven Material Defect Detection] {A Comprehensive Survey on Machine Learning Driven Material Defect Detection
}


\author{Jun Bai}
\affiliation{%
  \institution{Centre for Future Materials, University of Southern Queensland, Toowoomba}
   \country{Australia}
}
\email{jun.bai@unisq.edu.au}

\author{Di Wu}
\affiliation{%
  \institution{School of Mathematics, Physics
and Computing, University of Southern Queensland, Toowoomba}
   \country{Australia}
}
\email{di.wu@unisq.edu.au}

\author{Tristan Shelley}
\affiliation{%
  \institution{Centre for Future Materials, University of Southern Queensland, Toowoomba}
   \country{Australia}
}
\email{tristan.shelley@unisq.edu.au}

\author{Peter Schubel}
\affiliation{%
  \institution{Centre for Future Materials, University of Southern Queensland, Toowoomba}
   \country{Australia}
}
\email{peter.schubel@unisq.edu.au}

\author{David Twine}
\affiliation{%
  \institution{Centre for Future Materials, University of Southern Queensland, Toowoomba}
   \country{Australia}
}
\email{david.twine@unisq.edu.au}

\author{John Russell}
\affiliation{%
  \institution{Air Force  Research Laboratory, Wright-Patterson AFB OH}
   \country{USA}
}
\email{dr.john.d.russell@gmail.com}

\author{Xuesen Zeng}
\authornotemark[1]
\affiliation{%
  \institution{Centre for Future Materials, University of Southern Queensland, Toowoomba}
   \country{Australia}
}
\email{xuesen.zeng@unisq.edu.au}

\author{Ji Zhang}
\authornote{Corresponding authors.}
\affiliation{%
  \institution{School of Mathematics, Physics
and Computing, University of Southern Queensland, Toowoomba}
   \country{Australia}
}
\email{ji.zhang@unisq.edu.au}

\renewcommand{\shortauthors}{Jun Bai et al.}

\begin{abstract}
Material defects (MD) represent a primary challenge affecting product performance and giving rise to safety issues in related products. The rapid and accurate identification and localization of MD constitute crucial research endeavors in addressing contemporary challenges associated with MD.
In recent years, propelled by the swift advancement of machine learning (ML) technologies, particularly exemplified by deep learning, ML has swiftly emerged as the core technology and a prominent research direction for material defect detection (MDD). Through a comprehensive review of the latest literature, we systematically survey the ML techniques applied in MDD into five categories: unsupervised learning, supervised learning, semi-supervised learning, reinforcement learning, and generative learning. We provide a detailed analysis of the main principles and techniques used, together with the advantages and potential challenges associated with these techniques. Furthermore, the survey focuses on the techniques for defect detection in composite materials, which are important types of materials enjoying increasingly wide application in various industries such as aerospace, automotive, construction, and renewable energy. Finally, the survey explores potential future directions in MDD utilizing ML technologies. This survey consolidates ML-based MDD literature and provides a foundation for future research and practice.

\end{abstract}



\maketitle


\section{Introduction}

Material defect detection (MDD) stands as a critical subfield within the broader domain of defect detection, with a primary focus on identifying and assessing defects in various materials. It is integral to product quality assurance, production efficiency, and compliance with industry regulations, especially in manufacturing sectors where defects can compromise performance and safety \cite{usamentiaga2022automated}. MDD is a pivotal element in ensuring product reliability and safety, and it holds particular significance in the manufacturing industry. The potential impacts of defects necessitate robust detection methodologies, particularly given the increasing complexity of products and materials. Adherence to strict regulatory standards is crucial, not only to prevent catastrophic failures and reduce waste but also to maintain cost efficiency. In addition, MDD supports environmental sustainability by minimizing material waste and optimizing resource use, thus contributing to responsible production and innovation. Effective defect detection methodologies directly contribute to the longevity and performance of products, thereby driving innovation in manufacturing processes. This combination of quality assurance, safety, and sustainability underscores the critical role of MDD across diverse industrial applications.

Current MDD methods must adapt to the increasing complexity of manufacturing processes and product diversity to meet the demands of modern manufacturing. The challenges in MDD arise from the need to detect intricate defects, such as cracks, pores, and inclusions, which vary in shape, size, and location. One major challenge is the identification of complex defects. Additionally, the detection of minute defects, often imperceptible to the naked eye, requires advanced high-precision methods. Furthermore, MDD often involves handling multimodal data, including images, ultrasonic waves, and X-rays. Each data type has distinct features and noise levels, posing a challenge in effectively processing and integrating these diverse sources of information \cite{shi2020overview, ou2021recent}. Timely identification and resolution of issues during production demand real-time online detection with robust performance \cite{yang2020using, zeng2022small, zhao2023multi}. This requirement necessitates the development of methods with fast, stable, and reliable performance. Moreover, the rarity of defective samples and the significant resources required for defect labeling create additional barriers for efficient MDD \cite{jin2022survey, shahrabadi2022defect}.

Traditional approaches to MDD primarily rely on manual inspection, which faces inherent limitations in modern industrial applications. Manual inspection depends on human expertise, making it prone to subjectivity, fatigue, and inconsistency, particularly when detecting small or subtle defects \cite{bhatt2021image}. The increasing complexity of manufacturing processes and products further exposes the shortcomings of manual methods, as they often fail to deliver the precision and reliability required for identifying intricate or hidden defects. Additionally, the labor-intensive nature of manual inspection limits its scalability, making it impractical for high-speed or large-scale production lines. These constraints not only hinder the efficiency of quality control processes but also increase the likelihood of undetected defects, potentially compromising product performance and safety. To address these challenges, the industrial sector has increasingly turned to automated methods, which offer enhanced accuracy, speed, and scalability. This shift has paved the way for the adoption of machine vision and other advanced technologies in MDD, laying a foundation for more reliable defect detection systems.

As a response to the limitations of manual inspection, visual perception-based surface MDD has emerged as a highly effective alternative, progressively replacing traditional methods. Leveraging advancements in machine vision, these technologies enable non-contact, automated detection, providing higher accuracy and reliable performance even in complex production environments \cite{hafiz2021attention, ren2022state}. By integrating advanced image processing algorithms, visual perception-based systems are capable of automatically identifying and classifying defects, efficiently handling large data volumes, and performing high-precision inspections at industrial scale. These systems significantly enhance defect detection accuracy and consistency, minimizing reliance on human inspection and reducing the impact of human error, thereby ensuring more reliable and uniform inspection outcomes. The improved detection quality contributes directly to lowering production costs and boosting overall manufacturing efficiency \cite{ren2022state, luo2020automated, fan2020line}. The effectiveness of these technologies has been demonstrated across various sectors, including tile quality assurance \cite{stojanovic2001real}, fabric defect detection \cite{wang2021defect, dong2021automatic}, and steel plate inspection \cite{hu2020unsupervised, wang2018distributed}. Additionally, significant advancements have been achieved in PCB defect detection, as evidenced by the work of Zhang et al. \cite{zhang2021semi} and Mei et al. \cite{mei2018unsupervised}. By seamlessly integrating these capabilities, visual perception-based surface MDD systems address the challenges of modern manufacturing, setting new standards for quality assurance and operational efficiency.

Building on the advancements of vision-based MDD, machine learning (ML), a subset of artificial intelligence (AI), has further revolutionized automated MDD. ML augments the capabilities of machine vision by automating the entire detection process, from data preprocessing and feature extraction to defect identification and classification
\cite{nturambirwe2020machine}. Since its inception in the mid-20th century, ML has evolved from foundational concepts, such as early statistical methods, to algorithms designed to handle both structured and unstructured data, driven by advancements in computational power and the increasing availability of large datasets \cite{janiesch2021machine, cioffi2020artificial}. These advancements have enabled ML models to detect even the smallest defects with high precision, addressing challenges that are difficult for vision-based methods alone, such as subtle or hidden anomalies. ML ensures consistent performance, unaffected by human fatigue, and processes large data volumes rapidly, making it ideal for high-speed production environments. Beyond real-time applications, ML supports offline quality control by analyzing historical data to uncover defect trends and patterns, informing process improvements and reducing defect rates \cite{sresakoolchai2022railway, cao2020review}. By integrating ML with vision-based technologies, modern MDD systems achieve a new level of reliability and operational efficiency, setting benchmarks for quality assurance in manufacturing.

Despite the transformative potential of ML in MDD, several limitations hinder its broader adoption in industrial settings. One key challenge is the limited generalization ability of ML models, as detection algorithms are often tailored to specific materials or defect types, making them less effective when applied to new contexts or diverse manufacturing environments \cite{cui2021sddnet}. Additionally, ML-based systems require high-quality visual data for optimal performance, but industrial environments frequently present challenges such as complex backgrounds, noise, and inconsistent lighting conditions. The preparation of clean, high-resolution defect data is time-consuming and resource-intensive, posing barriers for seamless implementation. Maintenance and operational costs present further obstacles, as ML-driven systems depend on specialized hardware, advanced software, and trained personnel. The substantial initial investment, along with ongoing expenses for system upgrades, repairs, and training, can be particularly burdensome for small and medium-sized enterprises with constrained budgets \cite{caiazzo2022towards}. Addressing these limitations is essential to fully realize the potential of ML in MDD, enabling the development of robust, adaptable, and cost-effective solutions for diverse industrial applications.

Deep learning (DL) has emerged as a transformative ML approach in MDD, addressing limitations of traditional machine vision algorithms that often require task-specific designs. By constructing advanced network structures, DL enables the automatic extraction of high-level features from images, eliminating the need for manual feature engineering and streamlining defect detection and recognition processes \cite{hafiz2021attention}. This capability allows DL to effectively handle the diverse and complex requirements of contemporary MDD, including automation, high detection efficiency, accuracy, and real-time performance \cite{bhatt2021image}. The flexibility and adaptability of DL models make them suitable for a wide range of defect types and industrial settings, significantly advancing the field. Furthermore, the proliferation of high-performance computing resources, such as GPUs, has reduced the cost and complexity of deploying DL-based systems in MDD. These advancements have made DL not only a powerful tool for enhancing defect detection accuracy but also a practical and scalable solution for modern manufacturing challenges.

In this survey, while general ML-driven MDD techniques are extensively discussed, composite materials are highlighted separately due to their unique challenges. Unlike homogeneous materials such as metals or ceramics, composites exhibit anisotropy, heterogeneity, and complex failure mechanisms, which demand specialized defect detection methodologies \cite{P9,P17}. This distinction ensures a comprehensive exploration of both general approaches and those tailored to the specific needs of composite materials.

\begin{table}[!t]
\centering
\setlength{\abovecaptionskip}{5pt}
\setlength{\tabcolsep}{4pt}
\fontsize{8}{10}\selectfont
\caption{Comparison of our survey with related review papers on ML for Material Defect Detection}
\label{table:surs-cmp}
\begin{tabular}{l|c|c|c|c|c|c|c|c|c}
\hline
\multicolumn{1}{c|}{\textbf{Key Dimensions}}            & \cite{P2} & \cite{P3} & \cite{P5} & \cite{P6} & \cite{P7} & \cite{P8} & \cite{P9} & \cite{P10} & \textbf{Ours} \\ \hline
General ML Techniques for MDD            & \ding{55}   & \ding{51}   & \ding{51}   & \ding{51}   & \ding{55}   & \ding{51}   & \ding{51}   & \ding{51}    & \textbf{\ding{51}} \\ \hline
Focus on DL                   & \ding{51}   & \ding{55}   & \ding{55}   & \ding{51}   & \ding{51}   & \ding{51}   & \ding{55}   & \ding{51}    & \textbf{\ding{51}} \\ \hline
Specific Material Domains                & \ding{55}   & \ding{51}   & \ding{51}   & \ding{55}   & \ding{55}   & \ding{55}   & \ding{51}   & \ding{55}    & \textbf{\ding{51}} \\ \hline
Emerging Techniques (e.g., RL, SSL)      & \ding{55}   & \ding{55}   & \ding{55}   & \ding{55}   & \ding{55}   & \ding{55}   & \ding{55}   & \ding{55}    & \textbf{\ding{51}} \\ \hline
Cross-Domain Applicability               & \ding{55}   & \ding{55}   & \ding{55}   & \ding{55}   & \ding{55}   & \ding{55}   & \ding{55}   & \ding{55}    & \textbf{\ding{51}} \\ \hline
Emphasis on Industrial Deployment        & \ding{55}   & \ding{55}   & \ding{51}   & \ding{55}   & \ding{55}   & \ding{51}   & \ding{55}   & \ding{51}    & \textbf{\ding{51}} \\ \hline
Comprehensive ML Framework for MDD      & \ding{55}   & \ding{55}   & \ding{55}   & \ding{55}   & \ding{55}   & \ding{55}   & \ding{55}   & \ding{55}    & \textbf{\ding{51}} \\ \hline
\end{tabular}
\vspace{-0.45cm}
\end{table}

The existing literature on defect detection predominantly focuses on general methodologies or specific application domains \cite{P2, P6, P7, P8, P10, P11}, with some studies delving into defect detection in particular types of materials \cite{P3, P5, P9, P17}. However, a systematic review of ML-driven MDD remains largely unexplored. As shown in Table~\ref{table:surs-cmp}, our survey addresses key gaps in the existing literature by providing a comprehensive framework and emphasizing emerging techniques. Our survey presents several distinct advantages that highlight its contributions to the field:
\begin{enumerate}
\item Firstly, we systematically gather the most recent and relevant literature regarding ML-driven MDD, thereby ensuring a comprehensive and methodical coverage of the subject matter. Our survey encompasses a broad spectrum of research works, providing a holistic view of the field;
\item Furthermore, we classified the various ML technologies currently employed in MDD applications. This classification is followed by a thorough analysis of each technology category. Such a comprehensive evaluation aids in discerning the suitability and effectiveness of different technological approaches in various contexts;
\item Moreover, our survey places special emphasis on MDD techniques specifically designed for composite materials. Composite materials are gaining increasing importance across various industries due to their unique properties and performance advantages. By dedicating attention to the challenges posed by the complex structures and heterogeneous nature of composite materials, we provide valuable insights into the state-of-the-art techniques tailored to address these specific issues;
\item Building on the analysis of existing technological works, we identified key challenges in current MDD practices and highlighted future directions of research in MDD. By providing these insights, we aim to guide researchers and practitioners toward addressing the limitations of current approaches and advancing the field of ML-driven MDD.
\end{enumerate}

The rest of this survey paper is organized as follows. In Section 2, we present the preliminaries and the problem formulation for MDD. Following this, in Section 3, we provide a detailed technical classification of ML techniques applied in MDD. We systematically analyze and compare the technical characteristics, application domains, and pros and cons of each technology category. Section 4 focuses on the ML techniques specifically designed for defect detection in composite materials. In Section 5, we outline the open issues in current MDD practices and highlight potential research directions for the future. Finally, we conclude our paper in Section 6, summarizing the key contributions of this comprehensive survey.

\section{Preliminaries and Problem Identification} 

\subsection{Acronyms and Their Definitions}

Our paper introduces a substantial number of technical and academic concepts. To ensure the presentation remains clear and concise, we utilize their respective acronyms throughout the paper. For the convenience of the reader, we have included a comprehensive table of acronyms, as shown in Table~\ref{table:acronyms}, which provides the corresponding definitions for all acronyms used in the paper. 

\begin{table}[ht]
\centering
\setlength{\tabcolsep}{4pt}
\fontsize{7}{8}\selectfont
\caption{Definitions of acronyms used in this study }
\label{table:acronyms}
 
\begin{tabular}{lll|lll|lll}
\hline
\textbf{Acronym} & \textbf{Definition} & & \textbf{Acronym} & \textbf{Definition} \\ \hline
AI & Artificial Intelligence & & APS & Aluminium Profile Surface \\ 
CAE & Convolutional Autoencoder && 
CCVAE & Conditional Convolutional Variational Autoencoder  \\
CMs & Composite Materials & & CNN & Convolutional Neural Network \\
DBN & Deep Belief Network & & DL & Deep Learning \\
ECT & Eddy Current Testing && GAN & Generative Adversarial Network \\
GL & Genetive Learning && GNNs & Graph Neural Networks \\
GFRP & Glass Fiber Reinforced Plastic & & HOG & Histogram of Oriented Gradients \\ 
HSI & Hyperspectral Imaging & & MD & Material Defect \\
MDD & Material Defect Detection  &&
 ML & Machine Learning \\
 MPT & Magnetic Particle Testing && NDT & Non-Destructive Testing \\
 PCA & Principal Component Analysis && PT & Penetrant Testing \\
  RL & Reinforcement Learning && RNNs & Recurrent Neural Networks\\
SVM & Support Vector Machine && SL & Supervised Learning\\ 
SSL & Semi-Supervised Learning && UL & Unsupervised Learning\\ \hline
\end{tabular}
\end{table}

\subsection{Defect Detection}

Defect detection technology primarily focuses on inspecting surface defects of test samples, such as spots, scratches, colour variations, and deficiencies, and obtains relevant information about the defects, including their location, type, size, contours, and other characteristic features \cite{ren2022state}. Previous defect detection heavily relied on manual methods, exhibiting drawbacks such as high costs, low efficiency, suboptimal accuracy, and the inability to conduct large-scale inspections. However, with the maturity of information perception sensor technologies, such as visual imaging techniques, the advancement of defect detection technologies based on ML has been propelled \cite{prunella2023deep, usamentiaga2022automated}. 

\begin{figure}[!t]
\centering 
\includegraphics[scale=.5]{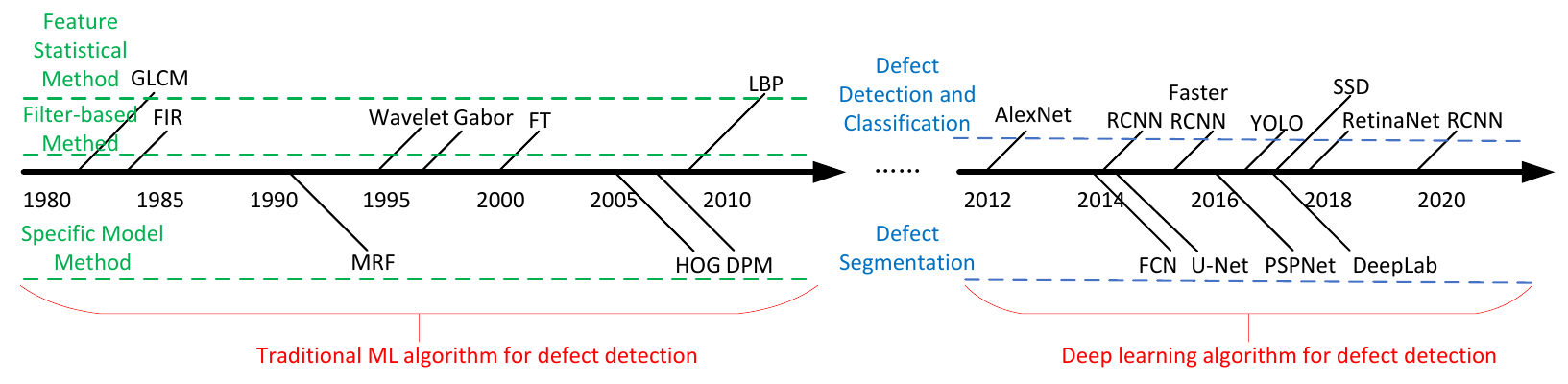} 
\centering 
\caption{Significant technical milestones of defect detection based on ML.}
 \label{fig:technical-milestones}
\vspace{-0.5cm}
\end{figure}

The present study delineates the developmental trajectory of defect detection technology based on ML techniques over time, as illustrated in Fig. \ref{fig:technical-milestones}. Notably, the delineation is anchored around the resurgence of neural networks in 2012, demarcating the overall progression of defect detection technology into two major phases: traditional ML algorithms and DL algorithms \cite{cao2020review}.

Traditional ML methods for defect detection include feature statistical methods, filter-based methods, and specific model methods. Statistical methods like Gray Level Co-occurrence Matrix (GLCM) \cite{zhu2015yarn} and Local Binary Pattern (LBP) \cite{tajeripour2007fabric} analyze spatial grayscale distributions to identify defects. Filter-based methods leverage signal processing to extract frequency domain features, utilizing techniques such as wavelet transform \cite{li2006improving}, Gabor filter \cite{kumar2002defect}, and Fourier transform \cite{chan2000fabric}. Specific models, such as Markov Random Field (MRF) \cite{dogandvzic2005defect} and Histogram of Oriented Gradients (HOG) \cite{halfawy2014automated}, are also employed for surface defect detection. These traditional methods excel in scenarios with limited defect types and low detection complexity, offering fast detection speeds.

With the resurgence of neural networks, DL has become a prominent area in defect detection \cite{bhatt2021image}. Two main DL categories stand out: defect detection/classification and defect segmentation. Detection/classification techniques, such as RCNN, Faster-RCNN, YOLO, and SSD, identify and localize defects while categorizing them \cite{tulbure2022review, qi2020review}. Segmentation algorithms, like U-Net and DeepLab, enable pixel-level classification but are more time-intensive \cite{tabernik2020segmentation}. Unlike traditional ML methods, DL algorithms automatically extract features and handle complex backgrounds and multi-defect scenarios effectively. However, they require significant computational resources and extensive annotated data.

\subsection{Material Defect Detection}

Material defect detection (MDD), as an important branch of defect detection, is a crucial area of research in industrial production to prevent safety issues caused by material defects. Rapid and non-destructive detection of potential defects in the material production process is a critical aspect of ensuring material quality. Common material defects can be categorized into surface defects and structural defects. Surface defects typically manifest as localized irregularities on the material surface, including irregular or abnormal features such as cracks, scratches, dents, colour variations, bubbles, wear, foreign objects, and more \cite{zhao2020hole, raizada2021surface}. Structural defects, on the other hand, involve abnormalities in the overall structure and shape of the material, such as shape defects (deformation, twisting, or bending), structural defects (improper connections, joints, or fixations), structural inconsistencies, and defects in connecting parts. Structural defects often exhibit a more complex manifestation in real-world applications and may include surface defects \cite{zhao2023structural, li2020strong, zhang2020defect}.

Common methods for material defect detection generally include magnetic particle testing (MPT), penetrant testing (PT), eddy current testing (ECT), ultrasonic testing, X-ray inspection, and machine vision inspection. 
Machine vision-based defect detection methods have significant advantages compared to other approaches. They enable automated non-destructive testing for virtually any material defect, offering high efficiency, safety, and precision. Importantly, defect detection through machine vision seamlessly integrates with the rapidly evolving field of AI, making existing material defect inspection more intelligent. This approach has garnered significant attention and preference from both academia and industry. Therefore, machine vision-based defect detection methods will serve as the central focus of this paper.

\begin{figure}[t]
\centering 
\includegraphics[scale=.37]{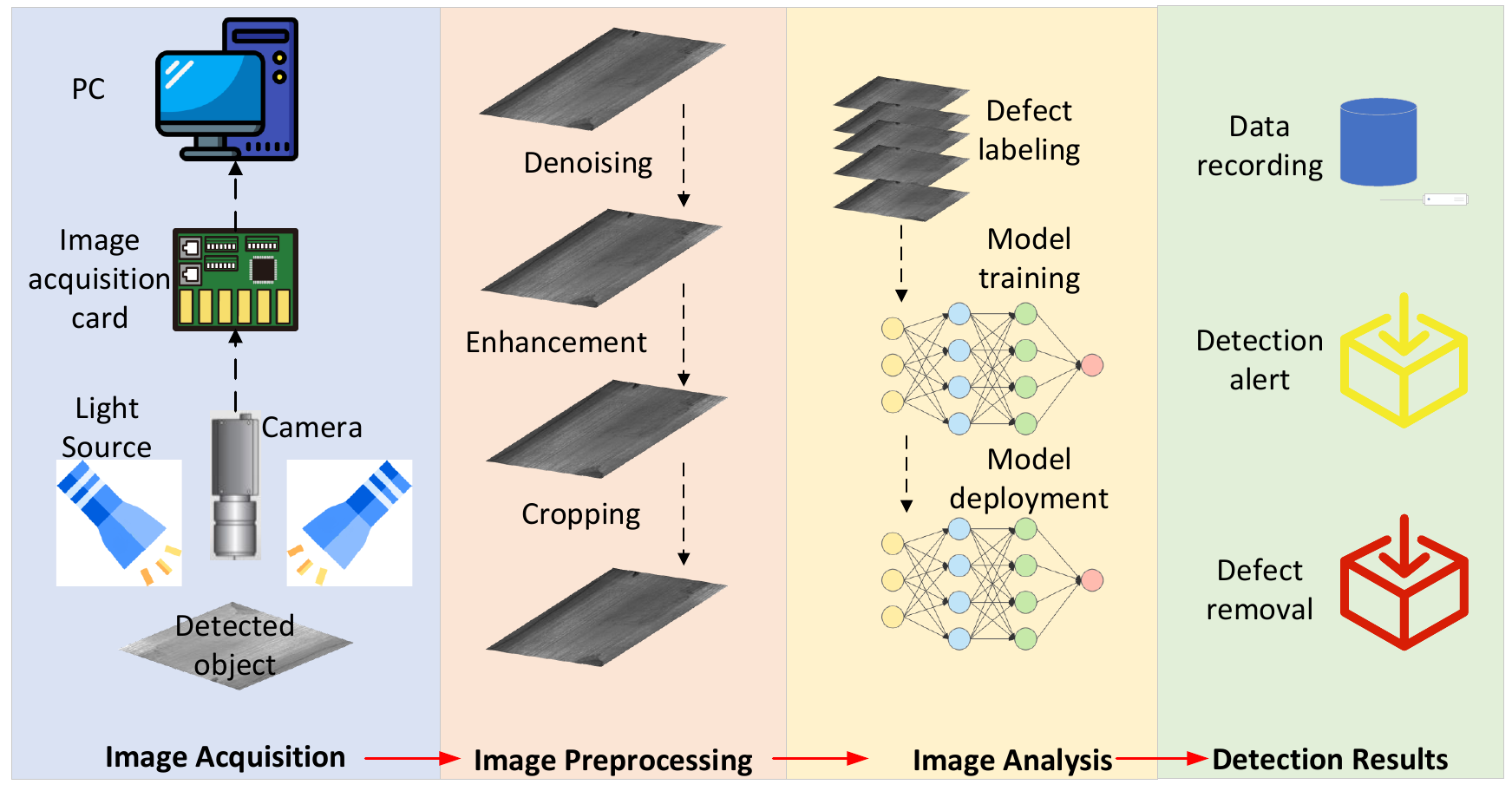} 
\centering 
\caption{Illustration of a material defect detection system based on machine vision.}
\label{fig:system}
\end{figure}

The machine vision defect detection system consists of four main components: image acquisition, image preprocessing, image analysis, and defect information output and recording, as illustrated in Fig. \ref{fig:system}. In the image acquisition phase, a visual system composed of lenses, lighting sources, industrial cameras, etc., is established to capture visual information of material defects in the form of image streams. After image acquisition, preliminary image preprocessing is required on the raw image information, including conventional operations such as noise reduction, image enhancement, and region of interest extraction, to prepare for subsequent image defect analysis.

The image analysis phase constitutes the core of the entire machine vision defect detection system, as specific defect detection algorithms will be deployed and executed here. For defect detection using traditional unsupervised ML algorithms, manually designed defect feature extraction and classification algorithms can be directly executed. If DL algorithms are employed for defect detection, steps such as defect image annotation, neural network model design, and model training are necessary, ultimately resulting in a defect detection model that can be deployed and executed. Lastly, the visual detection system records defect feature information outputted by the defect detection model, such as defect category, defect size, defect location, etc., in a database. Necessary actions are then taken for the defective material, such as alarms or removal.

\subsection{Experimental Evaluation and Benchmarks}

Benchmarking methods for MDD requires comprehensive experimental evaluations using standardized datasets and metrics to ensure reproducibility and enable fair comparisons across diverse defect types and complexities. Widely used datasets, such as NEU \cite{song2014automatic}, DAGM 2007 \cite{wieler2007weakly}, and Severstal \cite{severstal2019}, along with others listed in Table~\ref{table:datasets}, provide annotated defect images spanning various material categories, including steel and metal, fabric, texture, concrete, and industrial materials. These datasets facilitate targeted assessments, with NEU \cite{song2014automatic}, RSDDs \cite{Gan2017}, and UCI Steel Plates \cite{UCI_SteelPlatesFaults} focusing on steel defects, AITEX \cite{SilvestreBlanes2019} and Tianchi Cloth \cite{Qi2023} addressing fabric issues, and texture-related datasets like MVTec AD \cite{bergmann2019mvtec} and TILDA \cite{TILDA} covering surface patterns. Metrics such as accuracy, precision, recall, F1-score, and inference time are crucial for evaluating model performance, providing insights into overall effectiveness, robustness against errors, and suitability for real-time applications. By leveraging these datasets and metrics, researchers can perform standardized evaluations, fostering innovation, enabling fair comparisons, and identifying areas for improvement in defect detection methods.

\begin{table}[!t]
\centering
\fontsize{7.5pt}{10pt}\selectfont
\caption{Overview of common datasets for material defect detection}
\label{table:datasets}
\begin{tabular}{c|c|p{7cm}}
\hline
\textbf{\makecell[c]{Material Category}}             & \textbf{\makecell[c]{Dataset}}         & \textbf{\makecell[c]{Link}}         \\ \hline
\multirow{5}{*}{Steel and Metal} 
& NEU \cite{song2014automatic}              & http://faculty.neu.edu.cn/yunhyan/NEU\_surface\_defect\_ \newline database.html                              \\ \cline{2-3}
& RSDDs \cite{Gan2017}              & http://icn.bjtu.edu.cn/Visint/resources/RSDDs.aspx                                                \\ \cline{2-3}
& UCI Steel Plates \cite{UCI_SteelPlatesFaults}    & https://archive.ics.uci.edu/ml/datasets/Steel+Plates+Faults                                      \\ \cline{2-3}
& Severalst \cite{severstal2019}          & https://www.kaggle.com/c/severalst-steel-defect-detection/data                                    \\ \cline{2-3}
& GC10-Det \cite{Lv2020} & https://www.kaggle.com/datasets/alex000kim/gc10det                            \\ \hline

\multirow{3}{*}{Fabric}       
& AITEX \cite{SilvestreBlanes2019}              & https://www.aitex.es/afid/                                                                         \\ \cline{2-3}
& Fabric Stain \cite{FabricStainDataset2023}       & https://www.kaggle.com/priemshpathirana/fabric-stain-dataset                                      \\ \cline{2-3}
& Tianchi Cloth \cite{Qi2023}       & https://tianchi.aliyun.com/competition/entrance/ \newline 231666/information                               \\ \hline

\multirow{4}{*}{Texture and Miscellaneous}    
& TILDA   \cite{TILDA}            & https://hci.iwr.uni-heidelberg.de/node/3616                                                      \\ \cline{2-3}
& Kylberg  \cite{Kylberg2011}           & http://www.cb.uu.se/~gustaf/texture/                                                             \\ \cline{2-3}
& DAGM 2007  \cite{wieler2007weakly}         & https://hci.iwr.uni-heidelberg.de/node/3616                                                     \\ \cline{2-3}
& MVTec AD  \cite{bergmann2019mvtec}          & https://www.mvtec.com/company/research/datasets/mvtec-ad                                        \\ \hline

\multirow{3}{*}{Concrete and Building}        
& SDNET 2018  \cite{Dorafshan2018}        & https://www.kaggle.com/aniruddhsharma/structural-defects \newline -network-concrete-crack-images           \\ \cline{2-3}
& Surface Crack \cite{Ozgenel2018}       & https://www.kaggle.com/arunrk7/surface-crack-detection                                           \\ \cline{2-3}
& Bridge Cracks  \cite{Li2019}   & https://github.com/lsyksjr/Bridge\_Crack\_Image\_Data                                            \\ \hline

\multirow{4}{*}{Industrial}       
& Kolektor \cite{Tabernik2020}           & https://www.vicos.si/Downloads/KolektorSDD                                                      \\ \cline{2-3}
& DeepPCB  \cite{Tang2019}           & https://github.com/tangsanli5201/DeepPCB                                                        \\ \cline{2-3}
& elpv-dataset  \cite{Buerhop-Lutz2018}      & https://github.com/zae-bayern/elpv-dataset                                                      \\ \cline{2-3}
& NanoTWICE \cite{Carrera2017}          & https://www.mi.imati.cnr.it/ettore/NanoTWICE/                                                   \\ \hline
\end{tabular}
\vspace{-0.5cm}
\end{table}

\subsection{Challenges in Material Defect Detection} 

Compared to common defect detection tasks, MDD technology necessitates consideration of specific industry attributes, material characteristics, and the material production environment. Consequently, this introduces new challenges and complexities to the realm of MDD. The current primary challenges facing ML-based MDD include:

\begin{itemize}

    \item \textbf{Sample Scarcity:} Sample scarcity is a common challenge faced by all current defect detection systems, primarily manifested in three aspects: (1) With the advancement of modern production processes, strict control over the yield of substandard products has become a norm in current material production. This results in a scarcity of defect samples compared to readily available normal samples. (2) The collection and annotation of defect sample data incur high costs, rendering the acquisition of high-quality defect samples even more challenging. (3) Due to factors such as business competition and privacy protection, defect samples are rarely disclosed publicly, making it difficult to access actual defect samples.

  \item \textbf{Identification of Complex Defects:} Recognizing complex defects poses a significant challenge in ML for MDD. Efficient feature extraction and classification algorithms are required to accurately identify these defects, extracting useful features to enhance detection performance. The diverse types, shapes, sizes, and positions of defects in materials make identification challenging. Traditional detection methods often struggle to accurately identify complex defects such as cracks and pores.

   \item \textbf{Multimodal Data Processing:} Handling multimodal data is a crucial challenge in ML for MDD. Different data types exhibit distinct characteristics and noise levels, necessitating the development of specialized techniques to address these differences. Additionally, integrating data from various sources to improve detection performance presents a challenge. MDD involves multiple data types, including images, ultrasonic waves, X-rays, etc. Effectively processing these multimodal data, extracting useful features, and enhancing detection performance pose challenges.

   \item \textbf{Real-Time Online Detection:} Real-time online detection is a crucial challenge in ML for MDD. Developing efficient algorithms and optimizing models to reduce computation time and resource consumption are necessary to achieve real-time online detection. Maintaining model performance and accuracy in real-time environments is also challenging. Fast, reliable, and real-time defect detection is essential on production lines or in real-time applications, but some current ML methods may struggle to meet these requirements.

\end{itemize}

\subsection{Survey Overview} 

Facing numerous challenges in MDD, the rapid development of visual hardware and technologies, particularly those based on DL, has led ML-guided defect detection algorithms to gradually dominate research. Researchers have applied various ML algorithms to defect detection, significantly improving accuracy, efficiency, and applicability. This study, centred around defect detection technology, explores the application scenarios and advantages and disadvantages of different MDD technologies. It not only provides a comprehensive literature review and future research directions for subsequent researchers but also offers extensive guidance for industrial practitioners.

\begin{figure}[t]
\centering 
\includegraphics[scale=.48]{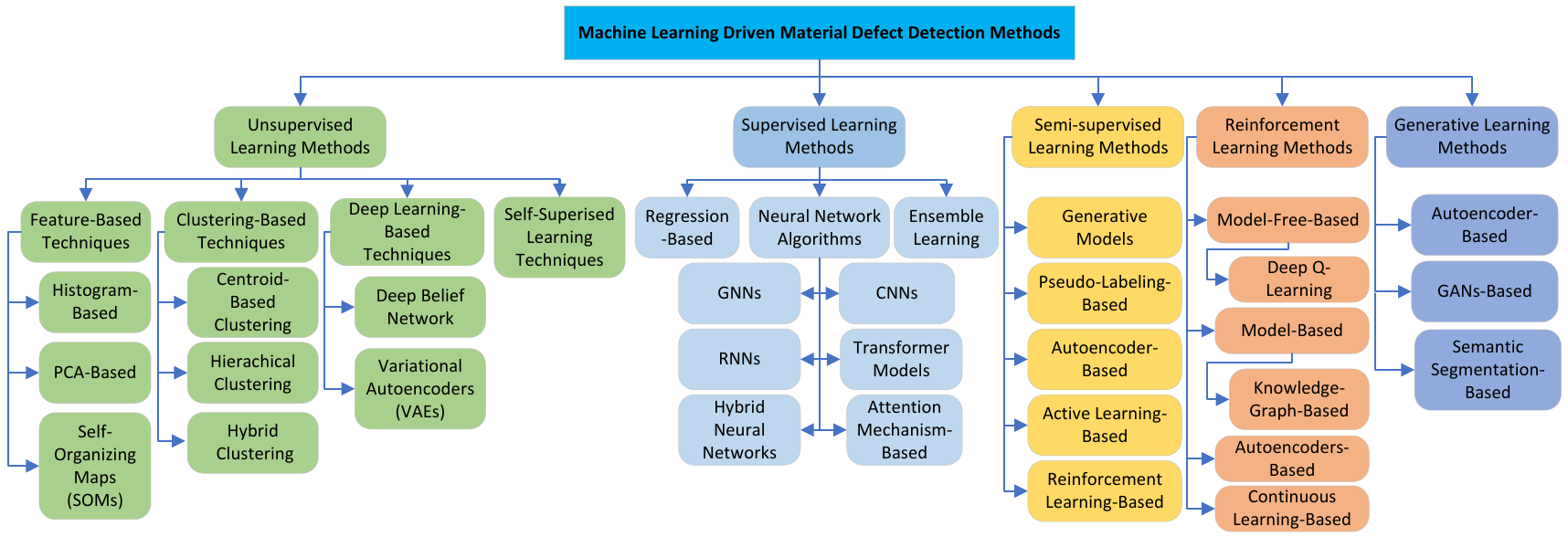} 
\centering 
\caption{Common machine learning methods for conducting the material defect detection.}
\label{fig:techniques}
\end{figure}

This paper summarizes recent research advancements in MDD, as illustrated in Fig. \ref{fig:techniques}, and provides detailed discussion and analysis in the subsequent sections. Our classification framework emphasizes the practical relevance and unique contributions of each technique in defect detection, categorizing them into unsupervised learning (UL), supervised learning (SL), semi-supervised learning (SSL), reinforcement learning (RL), and generative learning (GL). This distinction reflects the specialized roles these methods play in addressing diverse defect detection challenges.

UL algorithms are divided into traditional ML and DL methods. For scenarios where defect types are relatively simple and pre-defined, traditional unsupervised ML algorithms can utilize handcrafted features for rapid defect detection. For more complex cases without manual annotations, unsupervised DL methods are required to automatically learn features and distinguish defective samples from normal ones. GL, although often classified as a subset of UL, is treated separately in our framework due to its transformative role in addressing data scarcity. Techniques such as GANs and VAEs are instrumental in generating synthetic defect samples, balancing the sample distribution, and improving model robustness.

In situations where sufficient annotated samples are available, SL algorithms, particularly deep neural networks, can perform feature self-learning for defect detection, classification, and segmentation. However, given the scarcity of annotated samples in real-world industrial scenarios, semi-supervised or weakly SL methods are increasingly applied, enabling effective utilization of both labeled and unlabeled samples. RL is highlighted as a standalone category due to its unique application in dynamic and adaptive defect detection, particularly for real-time quality control and evolving industrial environments. 

Additionally, to meet industrial demands for real-time performance and high detection efficiency, model compression techniques are frequently applied during deployment to enhance operational efficiency. While ML-based defect detection has significantly improved detection accuracy and efficiency, challenges remain, including the complexity of defect types, limited generalization of detection models, and the high cost of model training and deployment. By categorizing methods based on their application-specific relevance and unique contributions, this framework aims to provide a clear, practical, and comprehensive perspective on material defect detection technologies.

\section{Techniques for Material Defect Detection } \label{sec:thematic_result}

\subsection{Unsupervised Learning Techniques in Material Defect Detection}

\begin{figure}[t]
\centering 
\includegraphics[scale=.42]{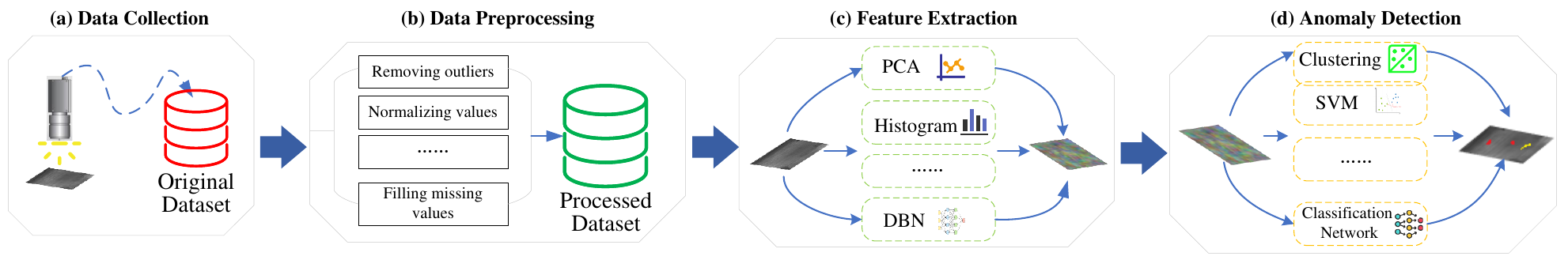} 
\caption{Illustration of the general working process for conducting the material defect detection with UL techniques.}
 \label{fig:ul-sys}
 \vspace{-0.5cm}
\end{figure}

This section elaborates on the UL techniques employed in MDD. Fig. \ref{fig:ul-sys} outlines its general workflow comprising four main steps: data collection, data preprocessing, feature extraction, and anomaly detection. Initially, a sensor captures data, forming the original dataset. This dataset is preprocessed to remove outliers, normalize values, and fill in missing data, producing a processed dataset suitable for subsequent analysis. Key features are extracted using methods such as Principal Component Analysis (PCA), histograms, and Deep Belief Networks (DBNs), which are categorized as feature-based techniques. Finally, anomaly detection is performed using techniques like clustering, DL models (e.g., VAEs and DBNs), and classification networks to identify and classify material defects.

\subsubsection{\textbf{Feature-Based Techniques}}

Feature-based techniques aim to extract meaningful and compact representations of raw data, emphasizing key characteristics that are critical for identifying defects. These methods transform complex, high-dimensional data into simplified forms, enabling subsequent detection algorithms, such as clustering or classification, to perform more efficiently and accurately. By highlighting essential features, feature-based techniques play a crucial role in enhancing the precision and robustness of defect detection workflows.

\paragraph{Histogram-Based}

Histogram-based techniques are widely applied in MDD \cite{kumar2021histogram}. These methods typically analyze texture features such as colour histograms and singular value decomposition (SVD) to identify defects effectively \cite{song2015wood}. For instance, Weighted Object Variance (WOV) enhances Otsu's method, improving segmentation accuracy by handling varying defect sizes and histogram distributions, resulting in higher detection rates and fewer false alarms \cite{yuan2015improved}. Similarly, Erazo-Aux et al. \cite{erazo2019histograms} introduce a method that combines histograms of oriented gradients with pulsed thermography for defect detection in composite materials, achieving robust results with an AUC score of 0.949. Additionally, Mahmoud et al. \cite{halfawy2014automated} employ a pattern recognition approach integrating HOG features and SVM classifiers for automating sewer defect detection from CCTV inspection images, demonstrating reliable segmentation and classification of defect areas.

\paragraph{Principal Component Analysis (PCA)-Based}

PCA is a statistical technique used in MDD to simplify the complexity of high-dimensional data while retaining trends and patterns \cite{cao2024crack}. Active thermography, a nondestructive testing method for composite materials, is improved in this study with a Sparse Principal Component Thermography (SPCT) method \cite{wu2018sparse}. Building on the advantages of Principal Component Thermography (PCT), SPCT introduces a penalization term, enhancing flexibility and interpretability due to its structural sparsity. The effectiveness of SPCT is proven through experiments on a carbon fibre-reinforced plastic specimen, demonstrating superior subsurface defect visualization compared to PCT. ML methods are increasingly significant in the field of nondestructive testing, particularly for the quality assessment of polymer composites. Among these, a Generative Adversarial Network (GAN), a form of DL, is introduced to thermography as an image augmentation tool, enhancing defect detection capabilities \cite{liu2020generative}. This method, termed GPCT, uses data augmentation to produce more diverse and informative images, improving the defect detection capabilities of thermographic analysis. The effectiveness of GPCT is demonstrated through successful experiments on a carbon fibre-reinforced polymer specimen, showcasing its potential in nondestructive testing.

\paragraph{Self-Organizing Maps (SOMs)-Based}

SOMs are a type of unsupervised neural network used for data visualization and pattern recognition, which can be effectively applied to MDD \cite{shakhovska2020improved}. An automated visual inspection method for detecting and classifying textile defects, using the autocorrelation function for feature extraction and a Self-Organizing Feature Map (SOFM) \cite{tolba1997self}. This approach is immune to illumination changes and noise, thanks to the noise rejection property of the autocorrelation function. Tested on real textile samples, the SOFM effectively clusters the feature vectors, showcasing the advantages of UL systems in defect classification. For textured pavement images, Mathavan et al. \cite{mathavan2015use} study a method using a SOM for detecting road cracks in textured pavement images. It involves analyzing images segmented into cells' texture and colour properties and using a Kohonen map to identify cracks. The technique achieves 77\% precision and 73\% recall, showing effectiveness even in uncontrolled environments and outperforming an advanced algorithm in reducing false positives in crack detection.

\subsubsection{\textbf{Clustering-Based Techniques}}

Clustering techniques group data into subsets based on feature similarities, making them effective for anomaly detection in material defects. These techniques can be broadly categorized into centroid-based, hierarchical, and hybrid clustering methods, each with unique strengths in MDD.

\paragraph{Centroid-Based Clustering (e.g., K-Means Clustering)-Based}

K-means clustering is widely applied in defect segmentation and clustering due to its simplicity and efficiency. For example, \cite{cohn2021unsupervised} achieved 99.4\% accuracy in classifying steel surface defects using features refined by PCA and extracted through a pre-trained VGG16 model. PCA reduced dimensionality, and K-means classified defects without requiring labeled data, offering valuable insights into unsupervised learning for materials science. Similarly, Thomas et al. \cite{leung2001representing} proposed a centroid-based clustering method for material recognition using 3D textons, compact surface patches encoding geometric and photometric properties. These textons, clustered into a vocabulary of representative patterns, enabled effective material recognition and appearance prediction under varied conditions.

\paragraph{Hierarchical Clustering-Based}

Hierarchical clustering constructs a hierarchy of clusters, enabling multi-scale defect detection. Jorge et al. \cite{igual2020hierarchical} present a hierarchical clustering approach using impact-echo techniques for defect detection in materials. Feature vectors, indicating wave propagation changes caused by defects, are analyzed as mixtures of Gaussians. Clusters are merged using the Kullback-Leibler divergence. Applied to aluminium alloy specimens, the method successfully classifies them into categories like no defect, one hole, one crack, and multiple defects, effectively distinguishing between defective and non-defective materials. Brzakovic et al. \cite{brzakovic1990approach} introduce an expert system for identifying and categorizing defects in materials with complex textures using digitized images. It involves a two-step image segmentation process and a material-specific defect classification approach. A hierarchical classification scheme incorporating shape, size, and texture descriptors is used for wood. The system, tested on parquet samples, has successfully detected and classified various defects.

\paragraph{Hybrid Clustering (e.g., Fuzzy and Spectral Clustering)-Based}
Hybrid clustering combines different clustering methods to handle more complex defect patterns. For example, in high-tech materials such as semiconductor manufacturing, a spatial defect diagnosis system has been proposed to estimate defect clusters and their separation, including both convex and non-convex types. The approach \cite{wang2008recognition} uses a spatial filter to clean wafer maps and a hybrid of entropy fuzzy c-means and spectral clustering to isolate defect patterns. A decision tree based on cluster features helps identify defect types. Tested with data from a DRAM company in Taiwan, the method effectively extracts and classifies various defect patterns, showing potential for application in other industries like liquid crystal and plasma display manufacturing.

\subsubsection{\textbf{Deep Learning (DL)-Based Techniques}}

DL-based techniques in UL are designed to automatically learn high-level feature representations from raw data, capturing intricate patterns and structures that traditional methods might overlook. These models, such as Variational Autoencoders (VAEs) and Deep Belief Networks (DBNs), excel in identifying complex and non-linear relationships within the data, making them particularly effective for anomaly detection and defect characterization in MDD workflows.

\paragraph{Variational Autoencoders (VAEs)-Based}

VAEs are a type of DL model particularly well-suited for anomaly and defect detection in materials. Using UL, Tsai et al. \cite{tsai2021autoencoder} evaluate an improved CAE model for automated defect detection in manufacturing. Unlike traditional CAEs that minimize reconstruction errors, this enhanced CAE incorporates a regularization to tighten the feature distribution of defect-free samples. This approach makes defect samples stand out more distinctly during evaluation. Tested on various material surfaces, including textured and patterned ones, the modified CAE demonstrates superior performance in defect detection compared to conventional CAEs. In addition, a VAE-enhanced DL model (VAEDLM) for efficiently classifying imbalanced wafer defects in semiconductor foundries \cite{wang2021variational}. The model effectively generates and recognizes wafer defect patterns by utilizing VAEs and a refined CNN. Tested on the WM-811K dataset, the VAEDLM achieves exceptional performance metrics, including 99.19\% accuracy and 99.96\% AUC, outperforming current methods. Additionally, the model's interpretability is enhanced through saliency maps and t-SNE, addressing a common limitation in DL applications.

\paragraph{Deep Belief Networks (DBNs)-Based}

A DBN is a type of deep neural network with multiple layers of stochastic, latent variables. It's particularly useful for feature extraction and classification tasks, making it suitable for MDD. Cang et al. \cite{cang2017microstructure} focus on enhancing Integrated Computational Materials Engineering (ICME) by introducing a Convolutional DBN (CDBN) for feature learning. This approach automates the transformation of microstructures into lower-dimensional representations, achieving a significant reduction in dimensionality. The model is applied to diverse material systems, including alloys and sandstone, successfully reconstructing materials closely resembling the originals in terms of key properties. This method outperforms traditional techniques that rely on the Markovian assumption, showcasing its potential in advanced material design and reconstruction. Furthermore, Liu et al. \cite{liu2019real} introduce a real-time quality monitoring and diagnosis scheme for manufacturing processes using a DBN. It addresses the challenge of handling complex, high-dimensional process data by converting it into quality spectra, which are then analyzed using a DBN. This model is developed through offline learning and applied to online process monitoring and diagnosis. Its effectiveness is demonstrated through simulation experiments, showing superior performance compared to traditional methods.

\subsubsection{\textbf{Self-Supervised Learning Techniques}}

Self-supervised learning (SSL) has emerged as a transformative approach within the realm of unsupervised techniques for MDD. SSL leverages unlabeled data by employing pretext tasks to generate pseudo-labels, enabling models to learn meaningful representations without manual annotations. This characteristic is particularly beneficial in scenarios where labeled defect data is scarce or expensive to obtain. For instance, Zhang et al. \cite{zhang2024maegan} proposed a Masked Autoencoder (MAE)-GAN model for cigarette defect detection, achieving robust performance in appearance defect identification. Similarly, Era et al. \cite{era2024thermal} employed SSL with Vision Transformers for melt pool characterization in directed energy deposition, enhancing thermal imaging-based defect detection. Anomaly detection in photovoltaic arrays using a dual-source SSL discriminator further demonstrates the method's versatility in handling noisy environments \cite{yao2024anomaly}. Additionally, Li et al. \cite{li2024steel} introduced a single-class SSL framework for surface defect detection in steel bars, effectively addressing imbalanced data distributions. Beyond these applications, Hu et al. \cite{hu2024segmentation} utilized SSL combined with multi-task learning to enhance defect segmentation accuracy, while Helwing et al. \cite{helwing2024damage} applied cycle-consistent GANs for damage evolution analysis in fiber-reinforced polymers. These advancements underscore SSL's ability to address complex defect scenarios by autonomously extracting and refining features, making it an indispensable tool for advancing MDD.

\subsubsection{\textbf{Unsupervised Technique Summary}}

\begin{table}[]
\setlength{\abovecaptionskip}{5pt}
\setlength{\tabcolsep}{4pt}
\fontsize{7.3}{10}\selectfont
\caption{Comparison of unsupervised learning techniques for material defect detection}
\label{table:uts}
\begin{tabular}{c|c|c|c}
\hline
\textbf{Technique Category}              & \textbf{Application Scenarios}                                                                                     & \textbf{Advantages}                                                                                                                                          & \textbf{Limitations}                                                                                                    \\ \hline
\textbf{Feature-Based Techniques}        & \begin{tabular}[c]{@{}c@{}}Specific defect types, \\ data simplification\end{tabular}                              & \begin{tabular}[c]{@{}c@{}}Simplifies high-dimensional \\ data; highlights key \\ features\end{tabular}                                                      & \begin{tabular}[c]{@{}c@{}}Limited to specific \\ feature extraction tasks\end{tabular}                                 \\ \hline
- Histogram-Based Techniques               & \begin{tabular}[c]{@{}c@{}}Detecting specific \\ defect types\end{tabular}                                         & \begin{tabular}[c]{@{}c@{}}Effective for targeted \\ defect identification\end{tabular}                                                                      & \begin{tabular}[c]{@{}c@{}}Not generalizable to \\ all defect types\end{tabular}                                        \\ \hline
- Principal Component Analysis (PCA)       & \begin{tabular}[c]{@{}c@{}}Preprocessing high-\\ dimensional data\end{tabular}                                     & \begin{tabular}[c]{@{}c@{}}Simplifies data; \\ retains key patterns\end{tabular}                                                                             & \begin{tabular}[c]{@{}c@{}}Not suitable for \\ complex defect types\end{tabular}                                        \\ \hline
- Self-Organizing Maps (SOMs)              & \begin{tabular}[c]{@{}c@{}}Pattern recognition \\ and data visualization\end{tabular}                              & \begin{tabular}[c]{@{}c@{}}Effective for clustering \\ feature vectors\end{tabular}                                                                          & \begin{tabular}[c]{@{}c@{}}Sensitive to parameters; \\ computationally intensive\end{tabular}                           \\ \hline
\textbf{Clustering-Based Techniques}     & \begin{tabular}[c]{@{}c@{}}Anomaly detection \\ and defect grouping\end{tabular}                                   & \begin{tabular}[c]{@{}c@{}}Groups similar data \\ points effectively\end{tabular}                                                                            & \begin{tabular}[c]{@{}c@{}}Requires significant \\ unlabeled data\end{tabular}                                          \\ \hline
- Centroid-Based Clustering (K-Means)      & \begin{tabular}[c]{@{}c@{}}Defect segmentation \\ and clustering\end{tabular}                                      & \begin{tabular}[c]{@{}c@{}}Simple and efficient for \\ clustering similar defects\end{tabular}                                                               & \begin{tabular}[c]{@{}c@{}}Less effective for \\ non-linear relationships\end{tabular}                                  \\ \hline
- Hierarchical Clustering                  & \begin{tabular}[c]{@{}c@{}}Multi-scale defect \\ detection\end{tabular}                                            & \begin{tabular}[c]{@{}c@{}}Handles multi-scale \\ defect patterns\end{tabular}                                                                               & \begin{tabular}[c]{@{}c@{}}High computational \\ complexity\end{tabular}                                                \\ \hline
- Hybrid Clustering                        & \begin{tabular}[c]{@{}c@{}}Complex defect \\ pattern detection\end{tabular}                                        & \begin{tabular}[c]{@{}c@{}}Combines strengths of \\ multiple clustering methods\end{tabular}                                                                 & \begin{tabular}[c]{@{}c@{}}Computationally \\ expensive\end{tabular}                                                    \\ \hline
\textbf{Deep Learning-Based Techniques}  & \begin{tabular}[c]{@{}c@{}}Anomaly detection \\ and defect characterization\end{tabular}                           & \begin{tabular}[c]{@{}c@{}}Automatically learns \\ high-level features\end{tabular}                                                                          & \begin{tabular}[c]{@{}c@{}}Requires extensive \\ training and data\end{tabular}                                         \\ \hline
- Variational Autoencoders (VAEs)          & \begin{tabular}[c]{@{}c@{}}Anomaly detection \\ in scarce data\end{tabular}                                        & \begin{tabular}[c]{@{}c@{}}Highlights anomalies; \\ generates diverse samples\end{tabular}                                                                   & \begin{tabular}[c]{@{}c@{}}Depends heavily on \\ training data\end{tabular}                                             \\ \hline
- Deep Belief Networks (DBNs)              & \begin{tabular}[c]{@{}c@{}}Feature extraction \\ and classification\end{tabular}                                   & \begin{tabular}[c]{@{}c@{}}Captures complex \\ defect features\end{tabular}                                                                                 & \begin{tabular}[c]{@{}c@{}}Computationally \\ intensive\end{tabular}                                                    \\ \hline
\textbf{Self-Supervised Learning (SSL)}           & \begin{tabular}[c]{@{}c@{}}Learning meaningful \\ representations from \\ unlabeled data\end{tabular}               & \begin{tabular}[c]{@{}c@{}}Reduces reliance on \\ labeled data; enhances \\ feature extraction\end{tabular}                                                  & \begin{tabular}[c]{@{}c@{}}Requires complex pretext \\ tasks and substantial \\ computational resources\end{tabular}     \\ \hline
\end{tabular}
\vspace{-0.5cm}
\end{table}

The application of UL techniques in MDD exhibits diversity and versatility. Table \ref{table:uts} summarizes the application scenarios, advantages, and limitations of various unsupervised techniques. As shown in Table \ref{table:uts}, feature-based techniques, including histogram-based methods, PCA, and SOMs, are commonly used for extracting key characteristics from raw data, enabling simplified representations for further analysis. Histogram-based techniques effectively detect specific types of defects, such as nodular defects on wood surfaces, but may not generalize to other defect types. PCA is valuable for reducing the complexity of high-dimensional data while retaining critical patterns, yet it is insufficient for addressing more complex defect scenarios. SOMs are useful for data visualization and clustering, showcasing their strength in pattern recognition, though their performance can be sensitive to parameter settings and environmental variability. Clustering-based techniques, such as K-Means, hierarchical clustering, and hybrid clustering, are highly effective for identifying defect patterns and grouping similar data points. These techniques offer broad applicability but often require significant amounts of unlabeled data for optimal performance. DL-based techniques, including VAEs and DBNs, excel in learning high-level representations and detecting anomalies or complex defect features. VAEs are particularly suitable for generating diverse samples and highlighting anomalies, making them effective in cases with scarce defect data, but they require extensive training data. DBNs are powerful tools for feature extraction and defect classification, but are computationally intensive and demand significant training resources. Additionally, SSL techniques have gained traction for leveraging unlabeled data through pretext tasks, enabling models to learn meaningful representations and improving defect detection in scenarios with limited labeled samples. Considering task requirements, data characteristics, and computational constraints, selecting the appropriate technique is crucial for improving the accuracy and efficiency of MDD.

\subsection{Supervised Learning Techniques in Material Defect Detection}

\begin{figure}[t]
\centering 
\includegraphics[scale=.4]{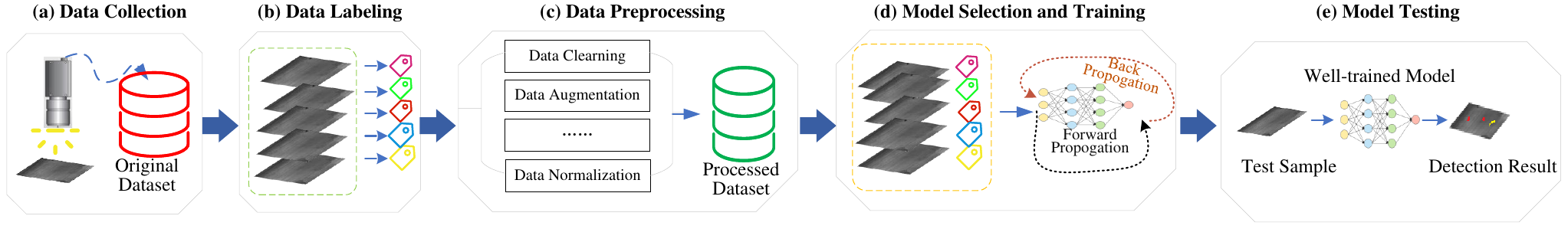} 
\centering 
\caption{Illustration of the general working process for conducting the material defect detection with supervised learning techniques.}
 \label{fig:sl-sys}
\vspace{-0.45cm}
\end{figure}

This section summarizes SL techniques relevant to MDD. Fig. \ref{fig:sl-sys} outlines a general workflow for this process, divided into five main steps: data collection, data labeling, data preprocessing, model selection and training, and model testing. Initially, a sensor captures data to form the original dataset. This dataset is then labeled with defect categories in the data labeling step. Next, the data undergoes preprocessing, which includes data cleaning, augmentation, and normalization, resulting in a processed dataset. In the model selection and training step, a designed learning model is trained using forward and backward propagation methods on the processed dataset. Finally, the well-trained model is tested with test samples to detect and classify material defects.

\subsubsection{\textbf{Regression-Based Techniques}}

Regression-based techniques are statistical and ML methods used for predicting continuous outcomes, such as defect depth, severity, or extent. These methods are commonly applied in scenarios requiring quantitative defect characterization. To detect the defect efficiently, a two-stage neural algorithm for defect detection and depth characterization using active thermography is proposed \cite{dudzik2015two}. The first stage involves a classification neural network for detecting defects, followed by a regression neural network to estimate their depth. Results from experimental investigations and simulations are presented, along with a sensitivity analysis that examines the effects of emissivity and ambient temperature errors on depth estimation. The algorithm's performance is validated using a test sample of low thermal diffusivity material. Various regression algorithms have been utilized in detecting defects during the Laser Beam Additive Manufacturing (LBAM) process. In their work, Mahmoudi et al. developed a system for detecting anomalies specifically in the Laser Powder Bed Fusion (LPBF) printing process \cite{mahmoudi2019layerwise}.

\subsubsection{\textbf{Neural Network Algorithms}}

Neural network algorithms play a critical role in MDD, leveraging diverse structures to address complex, nonlinear problems. These architectures are broadly categorized based on their design and application focus:

\paragraph{\textbf{Convolutional Neural Networks (CNNs)}}

CNNs are specialized for processing structured grid data, such as images, making them ideal for surface defect detection. Xing et al. \cite{xing2021convolutional} proposed an advanced CNN-based method for accurately detecting surface defects in workpieces. This model integrates a specialized feature extraction module and uses an optimized IOU-based loss function to enhance performance. The method demonstrated superior results over other models in both classification and detection tasks, achieving high accuracy and real-time detection capabilities. CNNs have thus become indispensable tools in industrial defect detection applications.

\paragraph{\textbf{Attention Mechanisms (Modules)-Based}}

Attention mechanisms, as modules integrated within neural network frameworks, enable models to focus on specific parts of the input data relevant to the defect detection task. These mechanisms significantly enhance the accuracy and efficiency of defect detection, particularly in challenging environments. For example, Fang et al. \cite{fang2022tactile} introduced a CNN-based method incorporating attention mechanisms for detecting fabric defects, using a vision-based tactile sensor. This approach improved detection robustness against ambient noise and complex dyeing patterns. Similarly, Chen et al. \cite{chen2021defect} utilized a self-attention mechanism for defect detection on aluminum profile surfaces, achieving superior performance in noisy conditions by refining feature extraction and employing a residual learning strategy.

\paragraph{\textbf{Transformer Models}}

Transformers, such as Vision Transformers (ViT), have emerged as powerful tools for defect detection by capturing global context and relationships in data. Liu et al. \cite{liu2024multi} proposed a Multi-feature Vision Transformer for automatic defect detection in composites, which integrates thermography with advanced feature representation. Wang et al. \cite{wang2023defect} developed Defect Transformer, a hybrid architecture combining CNN properties with transformer structures to enhance pixel-level defect detection precision in industrial materials like steel. These studies highlight transformers' ability to process high-resolution images and complex material characteristics.

\paragraph{\textbf{Graph Neural Networks (GNNs)}}

GNNs are increasingly used for analyzing structured data in material science. Pang and Tan \cite{pang2024steel} developed a steel surface defect detection model based on GNNs, leveraging graph representations to improve detection precision and efficiency. Thomas et al. \cite{thomas2023materials} applied GNNs for material fatigue prediction using microstructure representations, demonstrating their capability in capturing complex relationships within material microstructures. Additionally, Wang et al. \cite{wang2022graph} proposed a graph-guided CNN (G-GCNN) for surface defect recognition, which effectively combines graph-based representations and convolutional features to identify complex defect patterns. These findings showcase GNNs' versatility in modeling structured data and their effectiveness in industrial defect detection tasks.

\paragraph{\textbf{Recurrent Neural Networks (RNNs)}}

RNNs, particularly their variants such as Long Short-Term Memory (LSTM) networks, are effective in time-series defect detection, such as monitoring equipment performance or dynamic surface changes. These architectures capture temporal dependencies and trends critical for predictive maintenance. For example, Hu et al. \cite{hu2019lstm} applied LSTM-RNN to classify defects in honeycomb structures using infrared thermography, demonstrating high accuracy and efficiency in handling complex thermal data. Similarly, Wang et al. \cite{wang2020defect} proposed an LSTM-RNN-based method for determining defect depth in laser infrared thermography, achieving precise defect depth predictions.

\paragraph{\textbf{Hybrid Neural Networks}}

Hybrid neural networks integrate features from multiple architectures, such as CNNs, transformers, and attention mechanisms, to address complex defect detection tasks, especially in multimodal industrial scenarios. These networks leverage complementary strengths of different models to enhance detection precision and robustness. For instance, Zhao et al. \cite{zhao2024deep} developed a hybrid approach combining deep CNNs and multi-model fusion for optical inspection of hybrid bonded wafer defects. This method significantly improved detection reliability and accuracy for defects affecting hybrid bonding yields. Similarly, Zhou et al. \cite{zhou2023defect} introduced a hybrid attention network that integrates CNNs and auto-encoders (AE) for steel defect detection. Their model localized and classified defect regions with high precision, demonstrating robustness in challenging industrial conditions. These advancements underscore the potential of hybrid neural networks in achieving superior performance in diverse MDD applications.

\subsubsection{\textbf{Ensemble Learning-Based Techniques}}

Ensemble learning in MDD involves using multiple learning algorithms or models to obtain better predictive performance than could be obtained from any of the individual models alone \cite{alipour2020increasing}. A novel method for robotic hammering inspection aimed at maintaining social infrastructures, with a specific focus on estimating material defects like delamination depth in concrete \cite{fujii2016defect}. With the aging of infrastructure, there's a growing need for automated diagnostic methods. The hammering test, known for its efficiency in inspections, is gaining interest in robotic automation. The study introduces a method to estimate defect conditions suitable for detailed robotic inspections. It also incorporates an integration technique of multiple classifiers to enhance inspection accuracy. Additionally, the system is designed to minimize the impact of mechanical running noise. Experimental results with concrete test samples show the effectiveness of this approach, confirming its accuracy in both defect detection and condition estimation. In addition, a new deep ensemble learning method for real-time surface defect inspection in automated production focuses on overcoming challenges like domain shifts and imbalanced training data in flowing data scenarios \cite{liu2019data}. It introduces a unique distribution discrepancy identifier that blends deep CNNs with shallow learning for improved robustness and precision. The method was successfully tested on an industrial surface mount technology line, showing excellent mean average precision and adaptability for streaming data in industrial settings.

\begin{table}[!t]
\setlength{\abovecaptionskip}{5pt}
\setlength{\tabcolsep}{4pt}
\fontsize{7}{10}\selectfont
\caption{Comparison of supervised learning techniques for material defect detection}
\label{table:sts}
\begin{tabular}{l|l|l|l}
\hline
\textbf{\makecell[c]{Technique Category}}       & \textbf{\makecell[c]{Application Scenarios}}                                                                              & \textbf{\makecell[c]{Advantages}}                                                                                                     & \textbf{\makecell[c]{Limitations}}                                                                                   \\ \hline
\textbf{Regression-Based}         & \begin{tabular}[c]{@{}l@{}}Quantitative defect\\ characterization (e.g., \\ depth, size, severity)\end{tabular}  & \begin{tabular}[c]{@{}l@{}}Provides precise numerical \\ predictions and continuous \\ outcome estimation\end{tabular} & \begin{tabular}[c]{@{}l@{}}Highly dependent on labeled data \\ and limited generalizability \\ for unseen defect types\end{tabular} \\ \hline
\textbf{Neural Network}           & \begin{tabular}[c]{@{}l@{}}Diverse defect types \\ and complex, \\ nonlinear problems\end{tabular}                   & \begin{tabular}[c]{@{}l@{}}Adaptable to varied defect\\ scenarios; learns intricate\\ defect patterns\end{tabular}     & \begin{tabular}[c]{@{}l@{}}Requires large annotated \\ datasets; computational \\ overhead\end{tabular} \\ \hline
\makecell[c]{- Convolutional Neural \\ Network}   & \begin{tabular}[c]{@{}l@{}}Image-based defect \\ detection tasks\end{tabular}                                  & \begin{tabular}[c]{@{}l@{}}Robust texture analysis, \\ scalable to large datasets\end{tabular}                         & \begin{tabular}[c]{@{}l@{}}Sensitive to hyperparameter \\ selection; computationally \\ expensive\end{tabular} \\ \hline
\makecell[c]{- Recurrent Neural \\ Network}       & \begin{tabular}[c]{@{}l@{}}Time-series analysis \\ in defect detection\end{tabular}                             & \begin{tabular}[c]{@{}l@{}}Captures temporal patterns \\ for predictive maintenance\end{tabular}                       & \begin{tabular}[c]{@{}l@{}}Prone to vanishing gradient \\ issues; less suitable \\ for large-scale data\end{tabular} \\ \hline
- Transformer Models     & \begin{tabular}[c]{@{}l@{}}Global feature analysis \\ for high-resolution \\ defect detection\end{tabular}     & \begin{tabular}[c]{@{}l@{}}Captures global context; \\ excels in processing \\ structured data\end{tabular}            & \begin{tabular}[c]{@{}l@{}}Requires high computational \\ resources; needs large \\ datasets for training\end{tabular} \\ \hline
- Graph Neural Networks  & \begin{tabular}[c]{@{}l@{}}Relational defect \\ modeling (e.g., \\ microstructures)\end{tabular}               & \begin{tabular}[c]{@{}l@{}}Models complex relationships \\ in structured data\end{tabular}                             & \begin{tabular}[c]{@{}l@{}}Limited scalability; sensitive \\ to graph quality\end{tabular}             \\ \hline
    - Hybrid Neural Network        & \begin{tabular}[c]{@{}l@{}}Multimodal defect \\ detection combining \\ data types\end{tabular}                 & \begin{tabular}[c]{@{}l@{}}Combines multiple \\ architectures for improved \\ performance\end{tabular}                 & \begin{tabular}[c]{@{}l@{}}Increased model complexity \\ and resource requirements\end{tabular}        \\ \hline
\textbf{Attention Mechanisms}     & \begin{tabular}[c]{@{}l@{}}Localized defect detection \\ in complex environments\end{tabular}                   & \begin{tabular}[c]{@{}l@{}}Enhances focus on relevant \\ features, boosting accuracy\end{tabular}                      & \begin{tabular}[c]{@{}l@{}}Requires substantial computational \\ power; optimization \\ challenges\end{tabular}      \\ \hline
\textbf{Ensemble Learning}        & \begin{tabular}[c]{@{}l@{}}Combining models for \\ complex defect scenarios\end{tabular}                       & \begin{tabular}[c]{@{}l@{}}Improved robustness; \\ enhanced model \\ generalization\end{tabular}                       & \begin{tabular}[c]{@{}l@{}}High implementation complexity; \\ demanding on data \\ quality\end{tabular}               \\ \hline
\end{tabular}
\vspace{-0.45cm}
\end{table}

\subsubsection{\textbf{Supervised Technique Summary}}

This section summarizes the application of SL techniques in MDD, including regression-based techniques, neural network-based algorithms, and ensemble learning techniques, with a comparison provided in Table \ref{table:sts}. Regression-based techniques provide precise numerical predictions, making them suitable for quantitative defect characterization, such as depth, size, or severity, but they require highly annotated data and exhibit limited generalizability to unseen defect types. Neural network algorithms are applicable to complex and nonlinear problems, enabling accurate detection and quantitative assessment of defects, but they are further categorized into subtypes such as CNNs, RNNs, Transformer models, GNNs, and Hybrid Neural Networks, each offering unique advantages tailored to specific defect detection scenarios. For instance, CNNs excel in image-based tasks, while GNNs model structured relational data. These approaches demand extensive annotated datasets and incur significant computational costs. Attention mechanisms enable models to focus on task-relevant parts, enhancing the accuracy and efficiency of defect detection, particularly in localized and complex environments, but their integration increases computational and optimization complexity. Ensemble learning techniques improve predictive performance by combining multiple models, enhancing detection accuracy and robustness, although their high implementation complexity and reliance on high-quality defect data may pose challenges in practical deployment. Considering task requirements, data characteristics, and computational resources, selecting the appropriate techniques is crucial for improving the accuracy and efficiency of MDD.

\subsection{Semi-supervised Learning Algorithms}

In practical industrial scenarios, the collection and annotation costs of defective samples are often high, leading to a significant amount of unlabeled data. In order to make more effective use of these unlabeled data, researchers have turned to the setting of SSL \cite{liu2022defect, liu2021semisupervised, wang2021new}. In SSL, the training set includes both normal and abnormal samples, but only a small portion of the samples are labeled. Compared to SL, SSL reduces the dependence on labeled data.

In the latest works, \cite{di2019surface} proposes a SSL approach for the classification of surface defects in steel based on CAE and Semi-Supervised GAN (SGAN). This method begins by training a stacked CAE on a large amount of unlabeled data to extract features of surface defects in steel. After the completion of CAE training, the encoder network of CAE is retained as a feature extractor and connected to a softmax layer, forming a new classifier. Subsequently, SGAN is introduced for SSL to enhance the generalization capability of the new approach. The generator of SGAN can generate synthetic images of steel surface defects, which are used to train the classifier along with real images. The results indicate that CAE-SGAN achieves the best performance in defect classification, particularly for hot-rolled plates, with an approximately 16$\%$ increase in classification accuracy. \cite{he2019semi} presents a SSL approach for the classification of surface defects in steel. It utilizes a Conditional GAN for classification (cDCGAN) to perform UL on a limited set of annotated samples, generating a large number of unlabeled samples while simultaneously obtaining category predictions for each sample. Subsequently, unlabeled samples whose category predictions align with those of the classifier trained on a small set of labeled samples are added to the training set. Relevant experiments demonstrate that their method achieves high classification accuracy in the presence of limited labelled samples.

Similar to \cite{he2019semi}, \cite{gao2020semi} introduces an SSL approach based on CNN for steel surface defect identification. It leverages a small set of labeled samples and a large amount of unlabeled samples for model training. The model's generalization capability is enhanced by generating pseudo-labels for unlabeled samples, thereby using the pseudo-labeling technique. Also, the framework proposed by \cite{dai2020soldering} encompasses a comprehensive defect detection system, including both localization and classification of solder joints. For the classification of solder joints, a method integrating active learning and SSL is employed. Initially, an SVM classifier is trained on a small set of labelled data. During the training process, K-means clustering is utilized to analyze the distribution structure of the data, enabling the automatic or manual annotation of a portion of unlabeled samples. This expansion of the labeled dataset contributes to improving the performance of the classifier. Furthermore, \cite{chen2024weakly} proposes a weakly SL method leveraging multi-directional integrated circuit markings for surface defect inspection in surface mount technology. This approach employs a combination of weakly SL techniques and integrated circuit recognition to achieve high classification performance with minimal annotation requirements. Furthermore, to enhance defect detection performance, other learning methods are also integrated into SSL approaches. \cite{chu2020neural} proposes a semi-supervised framework for anomaly detection and segmentation to identify and segment abnormal regions in images. It employs an autoencoder to generate the reconstruction loss of images and combines it with an RL-based neural batch sampler to provide training batches for the autoencoder, thereby improving the performance of the predictor.

Compared to unsupervised methods, SSL significantly improves model performance when provided with a small amount of labeled defective samples. This approach is beneficial in addressing new types of defects in industrial product manufacturing. The application of SSL is primarily focused on defect classification at the image level, while research on defect localization tasks is still an area that requires further exploration.

\subsection{Reinforcement Learning Algorithms}

MDD often faces complex and dynamic environments, where the types and features of defects may exhibit diversity. Reinforcement learning (RL) can learn optimal strategies through interaction with the environment, enhancing the effectiveness of MDD. Therefore, RL, capable of meeting the demands of more complex material defect detection environments and achieving autonomous real-time dynamic learning of material defects, has started to gain favour among researchers.

 In \cite{chu2020neural}, an RL-based neural batch sampler optimizes training batches for an autoencoder, enhancing its ability to distinguish abnormal and normal regions through loss curves. Periodic reinitialization and retraining allow the autoencoder to optimize towards diverse local minima, improving prediction performance. Additionally, \cite{chung2022reinforcement} introduces Continuous G-Learning, an RL-based online defect detection method that leverages offline and online prior knowledge to adjust process parameters, ensuring rapid defect mitigation and improved quality assurance in additive manufacturing. The study also proposes an RL-based virtual texture surface defect generation system, where actor-critic networks search for optimal defect rendering parameters, generating realistic images that enhance training data diversity and detection recall rates. Furthermore, \cite{zhang2023duak} develops a bidirectional RL-based knowledge graph inference method for steel surface defect detection. This approach employs dual agents to explore paths in opposite directions, leveraging historical experiences to update policies, improving inference accuracy, and supplementing missing knowledge. Lastly, \cite{mohanty2024enhancing} integrates a Proximal Policy Optimization (PPO) algorithm within a digital twin model of titanium alloy washers, enabling an RL agent to simulate realistic defects for training a ResNet18-based detector, enhancing its robustness and real-world performance.

 RL methods can adapt to continuously changing environments in MDD. This is because they learn through interaction with the environment, enabling them to continuously adjust their strategies during training to accommodate new defect types or environmental changes. They also exhibit high real-time capability, allowing learning and decision-making in real-time data streams, which is crucial for scenarios requiring timely responses in MDD. Additionally, RL models can autonomously learn defect features without the need for extensive manual feature engineering. This enables them to extract valuable information from data and exhibit adaptability to complex defect situations. However, RL typically demands substantial training and computational resources, especially when dealing with complex material defect detection tasks. This may result in long training times and high computational costs. Moreover, RL models usually require a large number of samples for training, particularly in the case of deep RL models. Acquiring large-scale annotated data in the field of MDD can be an expensive and challenging task. Overall, RL holds potential in MDD.

\subsection{Generative Learning Algorithms}

In the industrial domain, the lack of precisely annotated defect samples makes it challenging to effectively train neural network models. To overcome this issue, data augmentation becomes a crucial factor in enhancing model performance. Common image augmentation techniques include cropping, rotation, distortion, and affine transformations \cite{zhou2023comparative, huang2020surface, dabetwar2022fatigue}. However, these methods, while altering the shape of defects, still maintain defects within the original background and may lack sufficient diversity, potentially leading to image distortion problems.

To obtain more diverse augmentation data, generative learning (GL) techniques are introduced into the augmentation of defect sample data, aiming to synthetically create realistic defects on normal samples. The work by \cite{valente2020print} introduces an end-to-end deep learning framework based on semantic segmentation for mapping printing defects. This framework enables pixel-level detection and localization of defects, such as stripes and colour bands, in printed images. Two types of printing defects and printing scan effects are simulated using image processing and computer graphics techniques, generating synthetic training datasets. This approach avoids the costs and time associated with manually collecting and annotating data. Experimental results demonstrate that the proposed framework can effectively learn features from synthetic data and generalize them to real printing defects. In \cite{bosse2023automated}, a data-driven feature labeling model is proposed for detecting hidden defects, such as pores in aluminium alloy die-cast plates, in low-cost X-ray projection images. Due to the absence of accurate annotated data for real X-ray images, the authors use numerical simulations of X-ray images to generate synthetic training datasets, taking into account various noise and distortion factors. Furthermore, \cite{farady2023preaugnet} introduces an offline PreAugmentation Network (PreAugNet) for generating and filtering augmented samples to improve the training of target networks. The PreAugmentation Network utilizes a CNN to extract features from original samples and updates the generated samples through the decision boundary analysis of an SVM classifier. This ensures that the new samples produced by the PreAugmentation Network are more diverse and representative. Evaluation on three real industrial steel surface defect datasets demonstrates the effectiveness of PreAugmentation Network data augmentation in improving the training of defect classification networks.

In addition to the methods mentioned above for generating defect data, the most common generative techniques are autoencoders and GANs. \cite{yun2020automated} proposes a metal surface defect classification algorithm using Deep Convolutional Neural Networks (DCNN) and a Conditional Convolutional VAE (CCVAE) for data augmentation. By learning the distribution of defect data, CCVAE generates diverse defect images, enhancing the DCNN model's generalization and accuracy. Similarly, GAN-based methods are widely utilized to generate defect images. \cite{pourreza2021g2d} introduces G2D (Generate to Detect Anomaly), leveraging GANs to produce anomaly samples during training, enabling anomaly detection in images and videos with only normal samples. \cite{zhang2021defect} presents Defect-GAN, which simulates damage and restoration processes to generate realistic defect samples, addressing the scarcity of defect data. Furthermore, \cite{jain2022synthetic} employs three GAN architectures (DCGAN, AC-GAN, and InfoGAN) to generate defect images for six types of steel strip defects. Their study explores latent variables in GANs to control synthetic image features like defect severity and lighting, incorporating domain knowledge and improving classification performance. These generative techniques demonstrate significant potential in augmenting datasets, particularly in scenarios with limited real-world defect samples.

In summary, GL methods can be employed in MDD to synthesize defect samples, thereby increasing the diversity of training data. This is particularly beneficial for MDD tasks that lack a large number of annotated defect samples, as it can significantly alleviate the cost of defect annotation, address data imbalance issues, and enhance model robustness. However, the performance of GL methods largely depends on the quality of the generated model, which may be constrained by the limitations of training data and may struggle to fully replicate the real-world distribution of defects. Additionally, synthesized defect samples may not be sufficiently diverse, leading to a decline in model performance in practical applications. Moreover, some GL methods, especially those based on DL, may have high computational resource requirements, especially when training and generating on a large scale. Therefore, the choice of GL methods should be comprehensively considered based on specific task requirements, data characteristics, and computational resources.

\subsection{Comparison for Diverse Material Defect Detection Tasks}

\begin{sidewaystable}[p]

\fontsize{12pt}{12pt}\selectfont
\vspace{32em}
\setlength{\extrarowheight}{2pt} 
\caption{Comparison for diverse material defect detection tasks.}
\label{tab:task-comparison}
\centering
\resizebox{\textwidth}{0.3\textwidth}{
\begin{tabular}{clll}
\hline
\textbf{Task Setting}    & \multicolumn{1}{c}{\textbf{Applicable Scenarios}}   & \multicolumn{1}{c}{\textbf{Advantages}}    & \multicolumn{1}{c}{\textbf{Limitations}}  \\ \hline

\begin{tabular}[c]{@{}l@{}} Unsupervised \\ Learning  \end{tabular}    
& \begin{tabular}[c]{@{}l@{}} $\bm{*}$ No labeled training data.\\ $\bm{*}$ Labeled samples of both defective \\ \quad and non-defective instances \\ \quad are scarce.\\ $\bm{*}$ Obvious defect features.\\ $\bm{*}$ Defects are easily distinguishable \\ \quad from the background.\\ $\bm{*}$ Relatively few types of defects.\end{tabular} 
& \begin{tabular}[c]{@{}l@{}} $\bm{*}$ Low development cost \\ \quad for the detection algorithm. \\ $\bm{*}$ High efficiency in real-time detection. \\ $\bm{*}$ Low cost for system setup \\ \quad and maintenance.\end{tabular}               
& \begin{tabular}[c]{@{}l@{}} $\bm{*}$ The model's performance is generally low. \\ $\bm{*}$ It cannot provide precise defect \\ \quad localization and classification. \\ $\bm{*}$ The model's robustness and \\ \quad generality are poor.\end{tabular} \\ \hline

\begin{tabular}[c]{@{}l@{}} Supervised \\ Learning   \end{tabular}    
& \begin{tabular}[c]{@{}l@{}} $\bm{*}$ Suitable for scenarios with ample \\ \quad annotated training data, including \\ \quad samples of both defects and non-defects.\\ $\bm{*}$ There are not too many restrictions on \\ \quad defect types and defect patterns.\end{tabular}  
& \begin{tabular}[c]{@{}l@{}} $\bm{*}$ High detection accuracy. \\ $\bm{*}$ Precise defect localization \\ \quad and classification. \\ $\bm{*}$ The model exhibits good robustness.\end{tabular}                        
& \begin{tabular}[c]{@{}l@{}} $\bm{*}$ Dependent on a large amount of \\ \quad labeled data. \\ $\bm{*}$ Difficult to adapt to new types of defects, \\ \quad requiring extensive re-labeling of data.\\ $\bm{*}$ Long development cycle for \\ \quad the detection model. \\ $\bm{*}$ High system setup and maintenance costs.\end{tabular} \\ \hline

\begin{tabular}[c]{@{}l@{}} Semi-supervised \\ Learning  \end{tabular} 
& \begin{tabular}[c]{@{}l@{}} Only a small fraction of the dataset \\ has labels, while the majority \\ of samples are unlabeled.\end{tabular}    
& \begin{tabular}[c]{@{}l@{}} $\bm{*}$ Utilizes unlabeled data to \\ \quad improve performance \\ \quad in situations where labeled data is scarce.\\ $\bm{*}$ Requires fewer labeled samples \\ \quad compared to SL.\end{tabular} 
& \begin{tabular}[c]{@{}l@{}} $\bm{*}$ Utilization of unlabeled data \\ \quad may not be fully effective, \\ \quad and performance remains constrained \\ \quad by the availability of labeled data.\\ $\bm{*}$ Quality of unlabeled data can \\ \quad impact model performance.\end{tabular} \\ \hline

\begin{tabular}[c]{@{}l@{}} Reinforcement \\ Learning \end{tabular}    
& \begin{tabular}[c]{@{}l@{}} $\bm{*}$ Applicable to scenarios that \\ \quad require autonomous learning of \\ \quad defect features. \\ $\bm{*}$ The defect detection system is capable of \\ \quad autonomously or semi-autonomously \\ \quad interacting with the inspection \\ \quad environment for learning.\end{tabular}  
& \begin{tabular}[c]{@{}l@{}} $\bm{*}$ Adapts to dynamic environments, \\ \quad offering advantages in real-time \\ \quad defect detection.\\ $\bm{*}$ Allows continuous improvement through \\ \quad interaction with the environment.\end{tabular} 
& \begin{tabular}[c]{@{}l@{}} $\bm{*}$ Requires extended training time, \\ \quad with high computational resource demands.\\ $\bm{*}$ Initial instability may be observed, \\ \quad necessitating prolonged training for stability.\end{tabular} \\ \hline

\begin{tabular}[c]{@{}l@{}} Generative \\ Learning    \end{tabular}    
& \begin{tabular}[c]{@{}l@{}} Applicable in situations where there are \\ relatively few defect samples or \\ styles of defect types, aiming to \\ increase the quantity of defect samples or \\ introduce variations in defect patterns.\end{tabular}    
& \begin{tabular}[c]{@{}l@{}} $\bm{*}$ Used for synthesizing defect data, \\ \quad enhancing the diversity of training data.\\ $\bm{*}$ Aids the model in learning defect \\ \quad features, improving robustness.\end{tabular}             
& \begin{tabular}[c]{@{}l@{}} $\bm{*}$ Training generative models itself may \\ \quad demand significant computational resources.\\ $\bm{*}$ The quality of generated data significantly \\ \quad influences the final model performance, \\ \quad requiring careful tuning of \\ \quad the generative model.\end{tabular} \\ \hline

\end{tabular}
}
\end{sidewaystable}

\paragraph{Performance Comparison Among Different ML Techniques}

In MDD, the use of different learning methodologies is crucial. Each methodology is designed for specific scenarios and has its own advantages and disadvantages. These are summarized in Table \ref{tab:task-comparison}, which provides a comparative analysis of the different approaches.
The choice of learning technique in MDD depends heavily on data availability, computational resources, and the complexity of the application context. UL provides flexibility in scenarios where labeled data is scarce, but its lack of explicit guidance often limits precision \cite{kumar2021histogram, yuan2015improved}. SL, leveraging methods like RCNN and YOLO, achieves high accuracy and precise defect localization but is constrained by the need for extensive labeled datasets and frequent updates to maintain relevance \cite{tabernik2020segmentation}. SSL bridges this gap by utilizing small labeled datasets alongside larger unlabeled ones, offering a cost-effective solution for complex and variable defect scenarios \cite{chu2020neural}. RL excels in dynamic environments through feedback-driven optimization, enabling advanced defect detection strategies but requiring significant computational resources and long training times \cite{chu2020neural, chung2022reinforcement}. Meanwhile, GL models, such as GANs and VAEs, enhance dataset diversity by synthesizing realistic defect samples, improving robustness and generalizability in data-scarce environments, though they are computationally intensive and require careful tuning \cite{zhang2021defect, jain2022synthetic}. Ultimately, while SL and SSL dominate in achieving accuracy and generalizability, RL and GL present innovative solutions for dynamic and data-augmented scenarios, balancing strengths and limitations to meet diverse industrial needs.

\begin{table}[!t]
\fontsize{8pt}{10pt}\selectfont
\caption{Performance comparison of traditional machine learning and deep learning techniques for material defect detection.}
\label{tab:comparison_ml}
\centering
\begin{tabular}{p{2.5cm}|p{5cm}|p{5cm}}
\hline
\textbf{\makecell[c]{Criteria}} & \textbf{\makecell[c]{Traditional ML Methods}} & \textbf{\makecell[c]{Deep Learning Methods}} \\ \hline
\textbf{\makecell[c]{Accuracy}} & Moderate accuracy, heavily dependent on handcrafted features. & High accuracy due to automatic feature extraction and end-to-end learning. \\ \hline
\textbf{\makecell[c]{Efficiency}} & Faster training and inference with low computational requirements. & Computationally intensive, requiring significant hardware resources for training. \\ \hline
\textbf{\makecell[c]{Robustness}} & Limited robustness to variations in defect types or environmental noise. & High robustness to defect variations, but may require large, diverse datasets for generalization. \\ \hline
\textbf{\makecell[c]{Feature Handling}} & Relies on manual feature engineering, which can be time-consuming and domain-specific. & Automatically extracts hierarchical features, reducing dependency on manual effort. \\ \hline
\textbf{\makecell[c]{Scalability}} & Effective for small datasets but struggles with scalability for complex problems. & Scales well to large datasets and complex defect detection tasks. \\ \hline
\textbf{\makecell[c]{Applicability}} & Suitable for simpler, well-defined problems with structured data. & Excels in complex, high-dimensional problems with diverse data modalities. \\ \hline
\textbf{\makecell[c]{Data Requirements}} & Performs well with limited labeled data. & Requires extensive labeled data for training to achieve optimal results. \\ \hline
\end{tabular}
\end{table}

\paragraph{Comparative Analysis of Traditional ML and DL Approaches}

Traditional ML methods, such as feature statistical techniques and filter-based approaches, have been widely applied in material defect detection due to their efficiency and simplicity. Techniques like the GLCM and Gabor filters excel in detecting well-defined defect types, offering fast processing with minimal computational demands. However, these methods rely on manually engineered features, which limit their adaptability to complex or heterogeneous defects, reducing their robustness in real-world scenarios.

In contrast, DL methods, including CNNs and Vision Transformers (ViTs), demonstrate significant improvements in accuracy and adaptability, particularly for intricate and overlapping defect types. By automatically extracting hierarchical features, DL models excel in diverse and challenging defect patterns. For example, CNNs are effective in detecting subtle surface defects, while ViTs capture global contextual information in high-resolution defect images \cite{liu2024multi}. Nevertheless, DL methods require substantial computational resources and large annotated datasets, making traditional ML approaches more practical in resource-constrained or simpler detection environments. Table \ref{tab:comparison_ml} summarizes these differences, highlighting the trade-offs between accuracy, efficiency, and robustness for both approaches.

\paragraph{Performance Comparison On Real-world Applications}

MDD is a cornerstone of industrial quality assurance, where ML techniques have proven to be transformative. Real-world applications, such as steel surface defect detection, fabric inspection, and concrete crack analysis, not only test the robustness of these techniques but also provide a foundation for benchmarking advancements. Across these applications, researchers have leveraged diverse datasets, such as NEU \cite{song2014automatic}, GC10-Det \cite{Lv2020}, MVTec AD \cite{bergmann2019mvtec}, and NanoTWICE \cite{Carrera2017}, to ensure standardized and fair evaluation. These benchmarks enable a direct comparison of traditional and modern approaches, highlighting how ML techniques have advanced the state of defect detection. 

As shown in Table~\ref{table:qpc-rwa}, advanced methods such as ViTs and YOLOv8 consistently outperform traditional techniques, achieving metrics like mAP scores above 98\% in steel surface \cite{song2014automatic} and textured surface defect detection \cite{bergmann2019mvtec}. Conversely, classical methods, such as histogram-based thresholding, exhibit limitations in handling complex scenarios, with lower F1-scores observed on datasets like SDNET 2018 \cite{Dorafshan2018}. This performance gap underscores the critical role of DL in adapting to diverse defect types. The adoption of such methods has not only improved accuracy and reliability but also established a roadmap for future innovations in industrial defect detection, ensuring methods are scalable and applicable across a wide range of domains.

\begin{table}[!t]
\centering
\fontsize{7.5pt}{10pt}\selectfont
\caption{Quantitative performance comparison on MMD real-world applications}
\label{table:qpc-rwa}
\begin{tabular}{p{1.9cm}|c|p{4.5cm}|c|c}
\hline
\textbf{\makecell[c]{Real-world \\ Application}}                        & \textbf{Ref.}                            & \textbf{\makecell[c]{Method}}                        & \textbf{Dataset (Ref.)}            & \textbf{\makecell[c]{Evaluation Metric \\ and Score}} \\ \hline
\multirow{4}{*}{\makecell[c]{Steel Surface \\ Defect Detection}} 
& \cite{song2014automatic}               & Scattering Convolution Network         & NEU \cite{song2014automatic}          & mAP: 96.3\%                   \\ \cline{2-5}
& \cite{cohn2021unsupervised}            & K-Means Clustering                     & GC10-Det \cite{Lv2020}                & Clustering Score: 91\%             \\ \cline{2-5}
& \cite{wang2024}                        & Improved YOLOv8                        & NEU-DET \cite{wang2024}              & mAP: 98.1\%                   \\ \cline{2-5}
& \cite{zhu2024mrp}                         & MRP-YOLO (Modified YOLOv8)             & NEU-DET \cite{wang2024}                & mAP: 95.4\%                        \\ \hline

\multirow{2}{*}{\makecell[c]{Fabric Defect \\ Detection}} 
& \cite{SilvestreBlanes2019}             & GLCM + SVM                             & AITEX \cite{SilvestreBlanes2019}      & F1-Score: 87\%                     \\ \cline{2-5}
& \cite{zhu2015autocorrelation}          & Autocorrelation + GLCM                 & Tianchi Cloth \cite{Qi2023}           & Precision: 88.5\%                  \\ \hline

\multirow{3}{*}{\makecell[c]{Concrete Crack \\ Detection}} 
& \cite{shalaby2024condition} & Histogram-based Threshold Detection & SDNET 2018 \cite{Dorafshan2018} & F1-Score: 74.11\% \\ \cline{2-5}
& \cite{Li2019}                          & Deep CNN for Crack Detection           & Surface Crack \cite{Ozgenel2018}      & mAP: 94.6\%                   \\ \cline{2-5}
& \cite{Dorafshan2018}                   & Deep CNN                               & SDNET 2018 \cite{Dorafshan2018}       & mAP: 92.3\%                   \\ \hline

\multirow{3}{*}{\makecell[c]{Texture Surface \\ Analysis}} 
& \cite{bergmann2019mvtec}               & Unsupervised Anomaly Detection         & MVTec AD \cite{bergmann2019mvtec}     & AUC: 95.4\%                        \\ \cline{2-5}
& \cite{li2021anomaly}                          & Self-organizing Map for Anomaly Detection (SOMAD) & MVTec AD \cite{bergmann2019mvtec}            & mAP: 97.9\%                     \\ \cline{2-5}
& \cite{tao2023vitalnet}                     & Vision Transformer & MVTec AD \cite{bergmann2019mvtec}            & mAP: 99.0\%                     \\ \cline{2-5}
& \cite{Lv2020}                         & Deep Learning-Based Feature Extraction & Kylberg \cite{Kylberg2011}            & Recall: 92.7\%                     \\ \hline

\multirow{2}{*}{\makecell[c]{Industrial Surface \\ Defects}} 
& \cite{Tabernik2020}                    & Segmentation-Based DL Method       & Kolektor \cite{Tabernik2020}          & Mean IoU: 93\%                     \\ \cline{2-5}
& \cite{Tang2019}                        & Group Pyramid Pooling (GPP)            & DeepPCB \cite{Tang2019}               & Precision: 90.8\%                  \\ \hline

\multirow{2}{*}{\makecell[c]{Miscellaneous \\ Applications}} 
& \cite{olimov2022uzadl} & Unsupervised Laplacian-based SOM & NanoTWICE \cite{Carrera2017} & mAP: 89.3\% \\ \cline{2-5}
& \cite{tao2023vitalnet}                     & Vision Transformer         & NanoTWICE \cite{Carrera2017}          & AUC: 94.5\%                        \\ \hline

\end{tabular}
\end{table}

\paragraph{Emerging Optimization Techniques for Real-Time Detection}

Recent advancements in neural architecture search (NAS) and model compression techniques have significantly improved the performance of ML models for real-time defect detection. NAS enables automated network design optimized for accuracy, latency, and resource constraints, making it suitable for industrial applications with stringent time requirements \cite{elsken2019neural, liu2021survey}. Additionally, techniques such as knowledge distillation have been effectively used to transfer the performance of large models to lightweight architectures. These smaller models, as demonstrated by Hinton et al. \cite{hinton2015distilling}, retain high accuracy while ensuring low latency for time-critical tasks.

Furthermore, hybrid edge-cloud computing frameworks have emerged as practical solutions for achieving high throughput and low latency in industrial settings. By processing time-sensitive tasks on edge devices and offloading computationally intensive operations to the cloud, these frameworks balance speed and scalability, as shown in steel surface and fabric defect detection systems \cite{xu2020edge, kang2017neurosurgeon}. Combined with hardware-specific optimizations and real-time processing pipelines, these advancements demonstrate the growing feasibility of deploying DL models in industrial environments where latency and throughput are critical.

\section{Techniques for Composite Material Defect Detection} 

Composite materials (CMs), widely used in high-tech fields such as aerospace and military due to their advantages of lightweight, high strength, and corrosion resistance, present unique challenges for defect detection. Unlike traditional homogeneous materials like metals, composites are anisotropic and heterogeneous, making them more prone to various types of defects, 
which can significantly degrade material performance or even lead to failure. Addressing these challenges requires advanced nondestructive evaluation (NDE) techniques and tailored ML models. While many general MDD algorithms discussed earlier are applicable, this section emphasizes adaptations and innovations specifically designed for CMs, including customized feature extraction methods and learning frameworks.

\subsection{Overview of Challenges in Composite Materials}
CMs require models capable of handling heterogeneous and complex data due to varied material properties and defect types, requiring advanced feature extraction and more complex models, such as DL architectures. Multi-modal data integration is often essential, combining information from various sensors, whereas other materials typically rely on single-mode data. Additionally, algorithms for CMs must be highly robust and adaptable to variations in defect presentation and environmental conditions. The inherent texture complexity in composites also emphasizes the need for texture analysis methods such as Gabor filters and GLCM, discussed later in this section.

\subsection{Texture Feature Analysis in Composite Materials}
Texture features play a vital role in detecting defects in composites due to their heterogeneous nature. Established methods like Gabor filters and GLCM have been widely utilized to extract and analyze texture features in composite defect detection \cite{osman2020automatic, hui2019application, li2018gabor, sharma2021grey, vasan2019defect}. Gabor filters are effective in capturing spatial frequency, orientation, and texture information, making them suitable for identifying localized surface defects in composites \cite{li2018gabor}. By applying multi-scale and multi-orientation Gabor filters, researchers have successfully extracted texture features that highlight defect patterns against complex composite backgrounds. GLCM, on the other hand, provides a statistical approach to analyzing texture by evaluating the spatial relationship of pixel intensities. Features derived from GLCM, such as contrast, correlation, energy, and homogeneity, offer a robust representation of composite material textures, enabling effective detection of defects like delamination or matrix cracking \cite{sharma2021grey, vasan2019defect}. Together, these methods enhance the precision of detecting defects in composites, addressing challenges posed by the materials' anisotropic and heterogeneous properties.

\subsection{Non-Destructive Testing Techniques}
NDT technology is one of the key methods for evaluating defects in CMs. Due to the heterogeneity and anisotropy of CMs, traditional metal material testing techniques are not always applicable. Common defect detection techniques for CMs include ultrasonic testing, X-ray testing, thermographic imaging, electronic speckle pattern interferometry, acoustic emission testing, and optical non-destructive testing. Below, we delve deeper into these methods based on their application domains.

\subsubsection{Ultrasonic Testing}
Ultrasonic testing is a widely used method for composite defect detection due to its ability to penetrate deep into materials and detect internal defects. For instance, \cite{mcknight20243} develops a DL approach using 3-D CNNs for detecting defects in carbon fibre-reinforced polymer composites using volumetric ultrasonic data. Meanwhile, \cite{oztacs2024image} examines the detection of internal damages in CMs using ultrasonic scanning, emphasizing its effectiveness in identifying defects like delamination. \cite{lu2024research} discusses enhancing ultrasonic testing with advanced signal processing techniques, which improve the detection of complex internal defects.

\subsubsection{X-Ray Testing}
X-ray testing provides high-resolution imaging for detecting internal defects in CMs. \cite{gong2020deep} proposes the use of deep transfer learning to enhance the detection of defects in X-ray images of aeronautic CMs, achieving a classification accuracy of 96\%.

\subsubsection{Thermographic Imaging}
Thermographic imaging has emerged as a powerful tool for detecting surface and subsurface defects in CMs. \cite{liu2024multi} presents a vision transformer model for thermography-based defect detection, while \cite{zhang2024one} introduces a one-dimensional deep CAE active infrared thermography (1D-DCAE-AIRT) approach to enhance defect visualization. Furthermore, \cite{deng2024attention} leverages a spatiotemporal-based DL model with an attention mechanism to analyze thermograms for detecting barely visible impact damages (BVID).

\subsubsection{Optical Non-Destructive Testing}
Optical methods are another promising approach for composite MDD. \cite{palka2024automatic} introduces a method using terahertz time-domain spectroscopy (TDS) to automatically detect defects in glass fibre-reinforced polymer (GFRP). Meanwhile, \cite{mahmoud2024enhancing} employs hyperspectral imaging (HSI) to detect defects in carbon fibre composites (CFC), enhancing detection speed and accuracy through spectral data analysis.

\subsubsection{Visualization for Composite Defects}

Visualization plays a critical role in composite MDD by providing interpretable representations of defects captured through various imaging techniques.
As presented in \cite{Vavilov2020},
multiple vision systems are employed to generate diverse defect visualizations. Optically-stimulated infrared thermography with different heating sources (tube or lamp heating) emphasizes surface and subsurface defects by creating high-contrast thermal maps. Ultrasonic techniques, including laser vibrometry and laser ultrasonics, enable precise imaging of internal defect structures by utilizing wave-based properties. These methods offer complementary perspectives on defect characterization, ranging from pixel-level defect identification to volumetric internal defect mapping. The integration of such visualizations into ML frameworks significantly enhances the interpretability and reliability of defect detection systems, providing a robust foundation for evaluating the structural integrity of CMs.


\subsection{Multimodal Data Integration for Composite Material Defect Detection}

The integration of multimodal data has emerged as a powerful approach for enhancing the accuracy and robustness of defect detection in CMs. By combining complementary data from different sources such as thermal imaging, ultrasonic testing, and optical sensing, multimodal integration addresses the limitations of single-modal methods, such as insufficient feature representation and sensitivity to specific environmental conditions. Recent advancements in this area have demonstrated the effectiveness of integrating data from multiple modalities to provide a more comprehensive understanding of defects in composite structures.

For instance, the fusion of phased-array ultrasonics (PAUT) and pulsed thermography (PT) has been successfully implemented using an Automatic Defect Recognition (M-ADR) system, enabling detailed structural evaluations of CMs. This multimodal approach not only improves the detection of internal defects like delamination but also enhances the resolution and reliability of surface defect assessments \cite{sudharsan2024multi}. Similarly, machine vision transformer networks (M-VIT-DDQ) have been developed to process multimodal thermal and optical data for defect detection in composites, leveraging multi-feature fusion to detect delamination and matrix cracking with high accuracy \cite{liu2024multi}. These techniques demonstrate the potential of combining multiple sensing modalities to achieve more accurate and versatile defect detection.

In addition to thermal and ultrasonic modalities, the synchronous fusion of optical and electromagnetic imaging has been explored to detect overlapping defects in complex composite geometries. By aligning and combining data from these modalities, researchers have achieved significant improvements in detecting subsurface defects that are challenging to identify using individual techniques \cite{ye2023synchronous}. Advanced ML models, such as DL frameworks and transformers, play a critical role in processing and interpreting multimodal inputs, enabling efficient feature extraction and integration of heterogeneous data.

Despite these advancements, challenges remain in the synchronization, dimensionality reduction, and noise suppression of multimodal data. To address these issues, recent studies have explored innovative solutions, such as tensor-based data analytics and DL-based fusion frameworks, which optimize the processing and integration of multimodal information. Moving forward, the integration of real-time processing capabilities and the development of domain-adaptive multimodal frameworks will be critical for advancing defect detection in CMs. By leveraging the strengths of multimodal data integration, researchers and practitioners can overcome the limitations of traditional single-modal approaches, achieving improved accuracy, efficiency, and reliability in composite MDD.

\section{Open Issues and Future Research Directions}
\label{sec::issues}

\subsection{Establishment of Standardized Datasets for Material Defects}

In the field of MDD, a significant challenge arises from the lack of standardized and widely accepted datasets. The absence of a standard dataset makes it difficult to achieve consistent evaluations of algorithms used in different studies. This hinders the comparability of research results and impedes the further development of the field. Additionally, the fragmented nature of available datasets and their limited coverage of diverse defect types pose challenges to the generalization ability of developed algorithms in practical applications.

To address these issues, it is crucial to focus on the development of standardized datasets that balance industrial needs, data privacy, and scalability. Therefore, it is important to create standardized and diverse MDD datasets that encompass various types of materials, multiple defect categories, and images with different scales and resolutions to ensure comprehensive testing of algorithms in various scenarios. For instance, CMs, which pose unique challenges due to their heterogeneous nature, should be specifically included to ensure datasets cover domain-specific scenarios. One potential solution is to encourage industry collaboration for the joint establishment of standardized datasets. By sharing data, research institutions and businesses can better assess their respective algorithms, promoting collaborative development across the entire field.  Moreover, data privacy and security concerns should be addressed through solutions such as federated learning and encryption techniques, allowing multiple stakeholders to contribute without exposing sensitive proprietary information. Additionally, advocating for the creation of open-source platforms enables researchers and practitioners to share algorithms, data, and experimental results, thereby advancing the open development of MDD technologies. Such platforms should include metadata standards and detailed annotations to ensure usability and reproducibility across different studies.

\subsection{Advanced Techniques for Complex Material Defect Detection}

Existing defect detection technologies usually focus on a singular type of defect. However, in real-world scenarios, materials often exhibit various types of defects simultaneously, and each type of defect may possess distinct shapes, sizes, and characteristics. Additionally, the interaction of multiple defects can further complicate detection tasks. For example, the presence of one defect may obscure the features of another defect. Furthermore, different types of defects may exhibit an imbalanced distribution in practical applications, potentially resulting in poor model performance on certain categories. This adds complexity and challenges to the task of detecting complex defects but is crucial for meeting the requirements of real-world MDD applications.

To address multi-category defect detection, a multi-task learning framework can be employed to simultaneously train models to handle multiple defect categories. This helps the model better understand the relationships between defects and improves overall performance. Furthermore, integrating multiple sensors or data sources, such as optical images, thermal imaging, acoustic waves, etc., can provide more comprehensive and multi-angle information, which helps enhance the ability to handle complex defects. 

\subsection{Small Sample Material Defect Detection}

In real-world applications of MDD, the advancement of manufacturing processes has led to a scarcity of defect samples compared to normal samples. Moreover, annotating defect samples requires a significant amount of human effort, further exacerbating the rarity of defect samples. This scenario is referred to as small sample MDD. Small sample MDD involves effectively detecting defects when there is an insufficient quantity of samples in the training data. However, most existing DL methods fall under the category of large sample learning and typically perform poorly in small sample situations. This results in models that struggle to accurately generalize to unseen defect samples, leading to low detection rates and unstable model performance, ultimately failing to meet practical inspection requirements.

Small sample defect detection poses a challenging problem due to the limitation of training data. To address this challenge, potential approaches include transfer learning, meta-learning, GANs, active learning, RL, and synthetic datasets. Transfer learning enhances a model's generalization ability in small sample contexts by leveraging knowledge pre-trained on other tasks. Meta-learning trains a model to handle multiple tasks, improving its adaptation to new defect detection tasks. GANs generate synthetic defect samples to augment training data. Active learning reduces annotation requirements by selecting samples with high information value. RL enhances performance through model interaction with the environment. Synthetic datasets create simulated defect data to increase training samples. The choice of these methods depends on the nature of the problem and available resources. Through their application, the performance and adaptability of small sample defect detection models can be improved, better meeting the diverse conditions encountered in actual industrial production.

\subsection{Dynamic Adaptation to Evolving Defect Patterns}

As manufacturing technology advances, the dynamic emergence of novel defects may pose challenges to existing defect detection methods within the current context. Firstly, the unknown shapes, features, or variations of novel defects make it difficult for existing models to accurately generalize to these changes. Secondly, the lack of standards or specifications for novel defects prevents existing detection methods from effectively adjusting based on prior experience. Additionally, the annotation data for novel defects may be extremely limited, leading to difficulties for existing SL methods in learning and adapting. Finally, addressing novel defects may require more real-time monitoring and detection, a demand that existing methods may not be able to meet.

To address the challenge of inapplicability in the current context, various potential solutions can be considered. Firstly, adopting incremental learning methods enables the model to update continually as it encounters new samples, adapting to the changes presented by novel defects. Secondly, utilizing adaptive algorithms allows the model to automatically adjust parameters based on the appearance of novel defects, enhancing adaptability to unknown defects. Transfer learning is another viable solution, transferring the model's experience from known defects to the detection of novel defects. Additionally, integrating multiple sensors or data sources, such as images, sounds, temperatures, etc., to form a multimodal fusion system can provide a more comprehensive, multi-angle perception of novel defects. By adopting these solutions, the capability of defect detection methods to dynamically adapt to novel defects can be enhanced, addressing the evolving demands of manufacturing technology.

\subsection{Real-time Online Defect Learning and Detection}

With the advancement of manufacturing technology, the demand for real-time defect learning and inspection continues to grow. Existing defect detection methods predominantly employ offline learning, wherein the model is trained using a pre-prepared dataset during the training phase and subsequently deployed to the production line. However, this approach faces limitations in the early stages of training due to a lack of annotated defect samples, making it challenging for the model to achieve optimal performance in actual deployment. Furthermore, offline learning methods are static by nature, making it difficult to accommodate dynamic changes in the production environment or variations in defect characteristics. In practical applications, production lines continually generate new samples and may introduce new types of defects, rendering offline learning methods less flexible. Therefore, employing adaptive online learning methods that can continuously integrate newly acquired data becomes a key issue in both research and application.

To address these potential issues, a series of solutions can be implemented. Firstly, incorporating incremental learning enables the model to be updated in real-time upon receiving new samples, enhancing real-time adaptability to novel defects. Incremental learning not only minimizes the risk of overfitting but also ensures the preservation of previously learned knowledge while accommodating new defect types. Secondly, optimizing algorithms and hardware infrastructure involves employing efficient computational methods and advanced hardware technologies to meet the computational requirements of real-time defect detection. Additionally, leveraging edge computing and distributed processing systems can significantly reduce latency, enabling faster response times for defect identification and correction. Stream data processing techniques can reduce the latency of sensor data reaching the detection system, improving real-time performance. Moreover, integrating anomaly detection mechanisms into the online learning pipeline can facilitate early detection of unknown defect patterns, providing an additional layer of robustness. Through the combined application of these solutions, it is possible to effectively address the issues of insufficient annotated data and the recognition of new defect types encountered by offline learning methods in practical applications. This makes the defect detection system more flexible, robust, and capable of real-time operation.

\subsection{Composite Material Defect Detection}

Composite materials are typically composed of multiple materials, which introduces heterogeneity and makes detection more challenging due to varying acoustical, thermal, and electrical properties among different constituents. Unlike homogeneous materials such as metals or ceramics, composites exhibit anisotropy and complex failure mechanisms, which often demand specialized ML-based defect detection methodologies. Furthermore, composites often exhibit complex geometries and structures, and defects may be located in areas that are difficult for sensors to access, thereby significantly increasing the difficulty of detection. Additionally, certain types of CMs possess high absorbance and scattering characteristics, introducing noise and interference to optical sensing devices, thus potentially reducing the accuracy of defect detection. Lastly, defects in CMs are often small and irregularly shaped, posing challenges for both detection and localization.

Comparatively, homogeneous materials such as metals typically have more uniform physical properties, which simplifies detection processes but still pose challenges in scenarios like detecting subsurface defects. This contrast underscores the importance of tailoring machine learning algorithms to the unique characteristics of composites. To address the aforementioned challenges in defect detection of CMs, future research efforts could focus on the comprehensive utilization of multimodal detection techniques, advanced imaging technologies, optimized sensor designs, and signal processing algorithms, as well as ML and AI approaches. These advanced technologies hold promise for improving the accuracy, efficiency, and reliability of defect detection in CMs.

\section{Conclusion}
\label{sec::summary}

In conclusion, this survey paper has demonstrated the increasingly vital role of ML techniques in the field of modern MDD. The integration of ML with machine vision technologies has led to significant advancements and practical applications in real-world defect detection scenarios. As technology continues to evolve, it is evident that ML-driven defect detection will remain a prominent area of research with far-reaching implications for various industries.

This paper has provided a comprehensive review of the latest research developments in MDD, covering task definitions, types of defects, challenges in detection, the framework of machine vision-based detection systems, and a detailed analysis of commonly used ML algorithms, including their respective advantages and disadvantages. By offering a thorough exploration of these aspects, this survey serves as a valuable resource for both researchers and practitioners in the field.

Looking ahead, the landscape of MDD is dynamic and continually evolving, presenting both challenges and opportunities. This paper has identified current research challenges and potential future directions, aiming to bridge the gap between theoretical research and industrial application. By highlighting these areas, we hope to inspire and guide future researchers, contributing to ongoing development and innovation in MDD. Furthermore, this survey provides practical insights that are invaluable for industrial practitioners. The comprehensive analysis of ML algorithms and their applications in defect detection can inform decision-making and strategic planning, enabling organizations to leverage the power of these technologies effectively. By understanding the capabilities and limitations of various approaches, practitioners can make informed choices and develop robust defect detection systems tailored to their specific needs.

\begin{acks}
This work is supported by the Centre for Future Materials at the University of Southern Queensland, Australia, and Air Force Research Laboratory (AFRL), USA.
\end{acks}

\bibliographystyle{unsrt}
\bibliography{MMD}

\begin{thebibliography}{100}

\bibitem{usamentiaga2022automated}
Rub{\'e}n Usamentiaga, Dario~G Lema, Oscar~D Pedrayes, and Daniel~F Garcia.
\newblock Automated surface defect detection in metals: A comparative review of object detection and semantic segmentation using deep learning.
\newblock {\em IEEE Transactions on Industry Applications}, 58(3):4203--4213, 2022.

\bibitem{shi2020overview}
Pengpeng Shi, Sanqing Su, and Zhenmao Chen.
\newblock Overview of researches on the nondestructive testing method of metal magnetic memory: Status and challenges.
\newblock {\em Journal of Nondestructive Evaluation}, 39:1--37, 2020.

\bibitem{ou2021recent}
Xiangyu Ou, Xue Chen, Xianning Xu, Lili Xie, Xiaofeng Chen, Zhongzhu Hong, Hua Bai, Xiaowang Liu, Qiushui Chen, Lin Li, et~al.
\newblock Recent development in x-ray imaging technology: Future and challenges.
\newblock {\em Research}, 2021.

\bibitem{yang2020using}
Jing Yang, Shaobo Li, Zheng Wang, Hao Dong, Jun Wang, and Shihao Tang.
\newblock Using deep learning to detect defects in manufacturing: a comprehensive survey and current challenges.
\newblock {\em Materials}, 13(24):5755, 2020.

\bibitem{zeng2022small}
Nianyin Zeng, Peishu Wu, Zidong Wang, Han Li, Weibo Liu, and Xiaohui Liu.
\newblock A small-sized object detection oriented multi-scale feature fusion approach with application to defect detection.
\newblock {\em IEEE Transactions on Instrumentation and Measurement}, 71:1--14, 2022.

\bibitem{zhao2023multi}
Zetian Zhao, Bingtao Hu, Yixiong Feng, Bin Zhao, Chen Yang, Zhaoxi Hong, and Jianrong Tan.
\newblock Multi-surface defect detection for universal joint bearings via multimodal feature and deep transfer learning.
\newblock {\em International Journal of Production Research}, 61(13):4402--4418, 2023.

\bibitem{jin2022survey}
Qifan Jin and Li~Chen.
\newblock A survey of surface defect detection of industrial products based on a small number of labeled data.
\newblock {\em arXiv preprint arXiv:2203.05733}, 2022.

\bibitem{shahrabadi2022defect}
Somayeh Shahrabadi, Yusbel Castilla, Miguel Guevara, Lu{\'\i}s~G Magalh{\~a}es, Dibet Gonzalez, and Telmo Ad{\~a}o.
\newblock Defect detection in the textile industry using image-based machine learning methods: a brief review.
\newblock In {\em Journal of Physics: Conference Series}, volume 2224, page 012010. IOP Publishing, 2022.

\bibitem{bhatt2021image}
Prahar~M Bhatt, Rishi~K Malhan, Pradeep Rajendran, Brual~C Shah, Shantanu Thakar, Yeo~Jung Yoon, and Satyandra~K Gupta.
\newblock Image-based surface defect detection using deep learning: A review.
\newblock {\em Journal of Computing and Information Science in Engineering}, 21(4):040801, 2021.

\bibitem{hafiz2021attention}
Abdul~Mueed Hafiz, Shabir~Ahmad Parah, and Rouf Ul~Alam Bhat.
\newblock Attention mechanisms and deep learning for machine vision: A survey of the state of the art.
\newblock {\em arXiv preprint arXiv:2106.07550}, 2021.

\bibitem{ren2022state}
Zhonghe Ren, Fengzhou Fang, Ning Yan, and You Wu.
\newblock State of the art in defect detection based on machine vision.
\newblock {\em International Journal of Precision Engineering and Manufacturing-Green Technology}, 9(2):661--691, 2022.

\bibitem{luo2020automated}
Qiwu Luo, Xiaoxin Fang, Li~Liu, Chunhua Yang, and Yichuang Sun.
\newblock Automated visual defect detection for flat steel surface: A survey.
\newblock {\em IEEE Transactions on Instrumentation and Measurement}, 69(3):626--644, 2020.

\bibitem{fan2020line}
Shuxiang Fan, Jiangbo Li, Yunhe Zhang, Xi~Tian, Qingyan Wang, Xin He, Chi Zhang, and Wenqian Huang.
\newblock On line detection of defective apples using computer vision system combined with deep learning methods.
\newblock {\em Journal of Food Engineering}, 286:110102, 2020.

\bibitem{stojanovic2001real}
Radovan Stojanovic, Panagiotis Mitropulos, Christos Koulamas, Yorgos Karayiannis, Stavros Koubias, and George Papadopoulos.
\newblock Real-time vision-based system for textile fabric inspection.
\newblock {\em Real-Time Imaging}, 7(6):507--518, 2001.

\bibitem{wang2021defect}
Yunyan Wang, Shuai Luo, and Huaxuan Wu.
\newblock Defect detection of solar cell based on data augmentation.
\newblock In {\em Journal of Physics: Conference Series}, volume 1952. IOP Publishing, 2021.

\bibitem{dong2021automatic}
Xinghui Dong, Chris~J Taylor, and Tim~F Cootes.
\newblock Automatic aerospace weld inspection using unsupervised local deep feature learning.
\newblock {\em Knowledge-Based Systems}, 221:106892, 2021.

\bibitem{hu2020unsupervised}
Guanghua Hu, Junfeng Huang, Qinghui Wang, Jingrong Li, Zhijia Xu, and Xingbiao Huang.
\newblock Unsupervised fabric defect detection based on a deep convolutional generative adversarial network.
\newblock {\em Textile Research Journal}, 90(3-4):247--270, 2020.

\bibitem{wang2018distributed}
Yalin Wang, Haibing Xia, Xiaofeng Yuan, Ling Li, and Bei Sun.
\newblock Distributed defect recognition on steel surfaces using an improved random forest algorithm with optimal multi-feature-set fusion.
\newblock {\em Multimedia Tools and Applications}, 77:16741--16770, 2018.

\bibitem{zhang2021semi}
Gaowei Zhang, Yue Pan, and Limao Zhang.
\newblock Semi-supervised learning with gan for automatic defect detection from images.
\newblock {\em Automation in Construction}, 128:103764, 2021.

\bibitem{mei2018unsupervised}
Shuang Mei, Hua Yang, and Zhouping Yin.
\newblock An unsupervised-learning-based approach for automated defect inspection on textured surfaces.
\newblock {\em IEEE Transactions on Instrumentation and Measurement}, 67(6):1266--1277, 2018.

\bibitem{nturambirwe2020machine}
Jean Frederic~Isingizwe Nturambirwe and Umezuruike~Linus Opara.
\newblock Machine learning applications to non-destructive defect detection in horticultural products.
\newblock {\em Biosystems engineering}, 189:60--83, 2020.

\bibitem{janiesch2021machine}
Christian Janiesch, Patrick Zschech, and Kai Heinrich.
\newblock Machine learning and deep learning.
\newblock {\em Electronic Markets}, 31(3):685--695, 2021.

\bibitem{cioffi2020artificial}
Raffaele Cioffi, Marta Travaglioni, Giuseppina Piscitelli, Antonella Petrillo, and Fabio De~Felice.
\newblock Artificial intelligence and machine learning applications in smart production: Progress, trends, and directions.
\newblock {\em Sustainability}, 12(2):492, 2020.

\bibitem{sresakoolchai2022railway}
Jessada Sresakoolchai and Sakdirat Kaewunruen.
\newblock Railway defect detection based on track geometry using supervised and unsupervised machine learning.
\newblock {\em Structural health monitoring}, 21(4):1757--1767, 2022.

\bibitem{cao2020review}
Wenming Cao, Qifan Liu, and Zhiquan He.
\newblock Review of pavement defect detection methods.
\newblock {\em Ieee Access}, 8:14531--14544, 2020.

\bibitem{cui2021sddnet}
Lisha Cui, Xiaoheng Jiang, Mingliang Xu, Wanqing Li, Pei Lv, and Bing Zhou.
\newblock Sddnet: A fast and accurate network for surface defect detection.
\newblock {\em IEEE Transactions on Instrumentation and Measurement}, 70:1--13, 2021.

\bibitem{caiazzo2022towards}
Bianca Caiazzo, Mario Di~Nardo, Teresa Murino, Alberto Petrillo, Gianluca Piccirillo, and Stefania Santini.
\newblock Towards zero defect manufacturing paradigm: A review of the state-of-the-art methods and open challenges.
\newblock {\em Computers in Industry}, 134:103548, 2022.

\bibitem{P9}
Sahar Hassani, Mohsen Mousavi, and Amir~H Gandomi.
\newblock Structural health monitoring in composite structures: A comprehensive review.
\newblock {\em Sensors}, 22(1):153, 2021.

\bibitem{P17}
Ronny~Francis Ribeiro~Junior and Guilherme~Ferreira Gomes.
\newblock On the use of machine learning for damage assessment in composite structures: A review.
\newblock {\em Applied Composite Materials}, pages 1--37, 2023.

\bibitem{P2}
Andrei-Alexandru Tulbure, Adrian-Alexandru Tulbure, and Eva-Henrietta Dulf.
\newblock A review on modern defect detection models using dcnns--deep convolutional neural networks.
\newblock {\em Journal of Advanced Research}, 35:33--48, 2022.

\bibitem{P3}
Aanchna Sharma, Tanmoy Mukhopadhyay, Sanjay~Mavinkere Rangappa, Suchart Siengchin, and Vinod Kushvaha.
\newblock Advances in computational intelligence of polymer composite materials: machine learning assisted modeling, analysis and design.
\newblock {\em Archives of Computational Methods in Engineering}, 29(5):3341--3385, 2022.

\bibitem{P5}
Yanzhou Fu, Austin~RJ Downey, Lang Yuan, Tianyu Zhang, Avery Pratt, and Yunusa Balogun.
\newblock Machine learning algorithms for defect detection in metal laser-based additive manufacturing: A review.
\newblock {\em Journal of Manufacturing Processes}, 75:693--710, 2022.

\bibitem{P6}
Zhonghe Ren, Fengzhou Fang, Ning Yan, and You Wu.
\newblock State of the art in defect detection based on machine vision.
\newblock {\em International Journal of Precision Engineering and Manufacturing-Green Technology}, 9(2):661--691, 2022.

\bibitem{P7}
Ryan Jacobs.
\newblock Deep learning object detection in materials science: Current state and future directions.
\newblock {\em Computational Materials Science}, 211:111527, 2022.

\bibitem{P8}
Jing Yang, Shaobo Li, Zheng Wang, Hao Dong, Jun Wang, and Shihao Tang.
\newblock Using deep learning to detect defects in manufacturing: a comprehensive survey and current challenges.
\newblock {\em Materials}, 13(24):5755, 2020.

\bibitem{P10}
Alireza Saberironaghi, Jing Ren, and Moustafa El-Gindy.
\newblock Defect detection methods for industrial products using deep learning techniques: A review.
\newblock {\em Algorithms}, 16(2):95, 2023.

\bibitem{P11}
Alireza Saberironaghi, Jing Ren, and Moustafa El-Gindy.
\newblock Defect detection methods for industrial products using deep learning techniques: A review.
\newblock {\em Algorithms}, 16(2):95, 2023.

\bibitem{prunella2023deep}
Michela Prunella, Roberto~Maria Scardigno, Domenico Buongiorno, Antonio Brunetti, Nicola Longo, Raffaele Carli, Mariagrazia Dotoli, and Vitoantonio Bevilacqua.
\newblock Deep learning for automatic vision-based recognition of industrial surface defects: a survey.
\newblock {\em IEEE Access}, 11:43370--43423, 2023.

\bibitem{zhu2015yarn}
Dandan Zhu, Ruru Pan, Weidong Gao, and Jie Zhang.
\newblock Yarn-dyed fabric defect detection based on autocorrelation function and glcm.
\newblock {\em Autex research journal}, 15(3):226--232, 2015.

\bibitem{tajeripour2007fabric}
Farshad Tajeripour, Ehsanollah Kabir, and Abbas Sheikhi.
\newblock Fabric defect detection using modified local binary patterns.
\newblock {\em EURASIP Journal on Advances in Signal Processing}, 2008:1--12, 2007.

\bibitem{li2006improving}
Xiaoli Li, Shiu~Kit Tso, Xin-Ping Guan, and Qian Huang.
\newblock Improving automatic detection of defects in castings by applying wavelet technique.
\newblock {\em IEEE Transactions on Industrial Electronics}, 53(6):1927--1934, 2006.

\bibitem{kumar2002defect}
Ajay Kumar and Grantham~KH Pang.
\newblock Defect detection in textured materials using optimized filters.
\newblock {\em IEEE Transactions on Systems, Man, and Cybernetics, Part B (Cybernetics)}, 32(5):553--570, 2002.

\bibitem{chan2000fabric}
Chi-ho Chan and Grantham~KH Pang.
\newblock Fabric defect detection by fourier analysis.
\newblock {\em IEEE transactions on Industry Applications}, 36(5):1267--1276, 2000.

\bibitem{dogandvzic2005defect}
Aleksandar Dogand{\v{z}}i{\'c}, Nawanat Eua-anant, and Benhong Zhang.
\newblock Defect detection using hidden markov random fields.
\newblock In {\em AIP conference proceedings}, volume 760, pages 704--711. American Institute of Physics, 2005.

\bibitem{halfawy2014automated}
Mahmoud~R Halfawy and Jantira Hengmeechai.
\newblock Automated defect detection in sewer closed circuit television images using histograms of oriented gradients and support vector machine.
\newblock {\em Automation in Construction}, 38:1--13, 2014.

\bibitem{tulbure2022review}
Andrei-Alexandru Tulbure, Adrian-Alexandru Tulbure, and Eva-Henrietta Dulf.
\newblock A review on modern defect detection models using dcnns--deep convolutional neural networks.
\newblock {\em Journal of Advanced Research}, 35:33--48, 2022.

\bibitem{qi2020review}
Shengxiang Qi, Jiarong Yang, and Zhenyi Zhong.
\newblock A review on industrial surface defect detection based on deep learning technology.
\newblock In {\em Proceedings of the 2020 3rd International Conference on Machine Learning and Machine Intelligence}, pages 24--30, 2020.

\bibitem{tabernik2020segmentation}
Domen Tabernik, Samo {\v{S}}ela, Jure Skvar{\v{c}}, and Danijel Sko{\v{c}}aj.
\newblock Segmentation-based deep-learning approach for surface-defect detection.
\newblock {\em Journal of Intelligent Manufacturing}, 31(3):759--776, 2020.

\bibitem{zhao2020hole}
Xiaoming Zhao, Chao Yao, Kaichen Gu, Tianran Liu, Yu~Xia, and Yueh-Lin Loo.
\newblock A hole-transport material that also passivates perovskite surface defects for solar cells with improved efficiency and stability.
\newblock {\em Energy \& Environmental Science}, 13(11):4334--4343, 2020.

\bibitem{raizada2021surface}
Pankaj Raizada, Vatika Soni, Abhinandan Kumar, Pardeep Singh, Aftab Aslam~Parwaz Khan, Abdullah~M Asiri, Vijay~Kumar Thakur, and Van-Huy Nguyen.
\newblock Surface defect engineering of metal oxides photocatalyst for energy application and water treatment.
\newblock {\em Journal of Materiomics}, 7(2):388--418, 2021.

\bibitem{zhao2023structural}
Biao Zhao, Yiqian Du, Zhikan Yan, Longjun Rao, Guanyu Chen, Mingyue Yuan, Liting Yang, Jincang Zhang, and Renchao Che.
\newblock Structural defects in phase-regulated high-entropy oxides toward superior microwave absorption properties.
\newblock {\em Advanced Functional Materials}, 33(1):2209924, 2023.

\bibitem{li2020strong}
Zhihan Li, Chaoji Chen, Ruiyu Mi, Wentao Gan, Jiaqi Dai, Miaolun Jiao, Hua Xie, Yonggang Yao, Shaoliang Xiao, and Liangbing Hu.
\newblock A strong, tough, and scalable structural material from fast-growing bamboo.
\newblock {\em Advanced Materials}, 32(10):1906308, 2020.

\bibitem{zhang2020defect}
Yiqiong Zhang, Li~Tao, Chao Xie, Dongdong Wang, Yuqin Zou, Ru~Chen, Yanyong Wang, Chuankun Jia, and Shuangyin Wang.
\newblock Defect engineering on electrode materials for rechargeable batteries.
\newblock {\em Advanced Materials}, 32(7):1905923, 2020.

\bibitem{song2014automatic}
Kechen Song, Shaopeng Hu, and Yunhui Yan.
\newblock Automatic recognition of surface defects on hot-rolled steel strip using scattering convolution network.
\newblock {\em J. Comput. Inf. Syst}, 10(7):3049--3055, 2014.

\bibitem{wieler2007weakly}
Matthias Wieler and Tobias Hahn.
\newblock Weakly supervised learning for industrial optical inspection.
\newblock In {\em DAGM symposium in}, volume~6, page~11, 2007.

\bibitem{severstal2019}
Kaggle.
\newblock Severstal: Steel defect detection dataset.
\newblock {\em Kaggle Competitions}, 2019.

\bibitem{Gan2017}
J.~Gan and Q.~J.~H. Li.
\newblock A hierarchical extractor-based visual rail surface inspection system.
\newblock {\em IEEE Sensors Journal}, 17(23):7935--7944, 2017.

\bibitem{UCI_SteelPlatesFaults}
Dheeru Dua and Casey Graff.
\newblock {UCI Machine Learning Repository: Steel Plates Faults Data Set}, 2019.

\bibitem{SilvestreBlanes2019}
J.~Silvestre-Blanes, T.~Albero-Albero, I.~Miralles, R.~Pérez-Llorens, and J.~Moreno.
\newblock A public fabric database for defect detection methods and results.
\newblock {\em Autex Research Journal}, 19:363--374, 2019.

\bibitem{Qi2023}
Y.~Qi, L.~Chu, D.~Yang, and Z.~Yu.
\newblock A clothing classification method based on residual network.
\newblock In {\em Third International Conference on Image Processing and Optical Engineering}. SPIE, 2023.

\bibitem{bergmann2019mvtec}
Paul Bergmann, Michael Fauser, Dominik Sattlegger, and Carsten Steger.
\newblock Mvtec ad–a comprehensive real-world dataset for unsupervised anomaly detection.
\newblock {\em Proceedings of the IEEE/CVF Conference on Computer Vision and Pattern Recognition}, pages 9592--9600, 2019.

\bibitem{TILDA}
{TILDA}.
\newblock Textile texture-database.
\newblock \url{http://lmb.informatik.uni-freiburg.de/resources/datasets/tilda.en.html}.

\bibitem{Lv2020}
X.~Lv, F.~Duan, J.~Jiang, X.~Fu, and L.~Gan.
\newblock Deep metallic surface defect detection: The new benchmark and detection network.
\newblock {\em Sensors}, 20:1562, 2020.

\bibitem{FabricStainDataset2023}
Priemsh Pathirana.
\newblock {Fabric Stain Dataset}.
\newblock \url{https://www.kaggle.com/datasets/priemshpathirana/fabric-stain-dataset}, 2023.

\bibitem{Kylberg2011}
G.~Kylberg.
\newblock The kylberg texture dataset, v. 1.0.
\newblock Technical Report Technical Report 35, Centre for Image Analysis, Swedish University of Agricultural Sciences, Uppsala, Sweden, 2011.

\bibitem{Dorafshan2018}
S.~Dorafshan, R.J. Thomas, and M.~Maguire.
\newblock Sdnet2018: An annotated image dataset for non-contact concrete crack detection using deep convolutional neural networks.
\newblock {\em Data in Brief}, 21:1664--1668, 2018.

\bibitem{Ozgenel2018}
Ç.F. Özgenel and A.G. Sorguç.
\newblock Performance comparison of pretrained convolutional neural networks on crack detection in buildings.
\newblock In {\em Proceedings of the International Symposium on Automation and Robotics in Construction (IAARC)}, pages 693--700, Berlin, Germany, 2018.

\bibitem{Li2019}
L.~Li, W.~Ma, L.~Li, and C.~Lu.
\newblock Research on detection algorithm for bridge cracks based on deep learning.
\newblock {\em Acta Automatica Sinica}, 45:1727--1742, 2019.

\bibitem{Tabernik2020}
D.~Tabernik, S.~Šela, J.~Skvarč, and D.~Skočaj.
\newblock Segmentation-based deep-learning approach for surface-defect detection.
\newblock {\em Journal of Intelligent Manufacturing}, 31:759--776, 2020.

\bibitem{Tang2019}
S.~Tang, F.~He, X.~Huang, and J.~Yang.
\newblock Online pcb defect detector on a new pcb defect dataset.
\newblock {\em arXiv preprint arXiv:1902.06197}, 2019.

\bibitem{Buerhop-Lutz2018}
C.~Buerhop-Lutz, S.~Deitsch, A.~Maier, F.~Gallwitz, S.~Berger, B.~Doll, J.~Hauch, C.~Camus, and C.J. Brabec.
\newblock A benchmark for visual identification of defective solar cells in electroluminescence imagery.
\newblock In {\em Proceedings of the 35th European PV Solar Energy Conference and Exhibition}, pages 1287--1289, Brussels, Belgium, 2018.

\bibitem{Carrera2017}
D.~Carrera, F.~Manganini, G.~Boracchi, and E.~Lanzarone.
\newblock Defect detection in sem images of nanofibrous materials.
\newblock {\em IEEE Transactions on Industrial Informatics}, 13:551--561, 2017.

\bibitem{kumar2021histogram}
Ranjeet Kumar, Anil~Kumar Soni, Aradhana Soni, and Saurav Gupta.
\newblock Histogram-based image enhancement and analysis for steel surface and defects images.
\newblock In {\em Machine Vision and Augmented Intelligence—Theory and Applications: Select Proceedings of MAI 2021}, pages 623--632. Springer, 2021.

\bibitem{song2015wood}
Weiwei Song, Tianyi Chen, Zhenghua Gu, Wen Gai, Weikai Huang, and Bin Wang.
\newblock Wood materials defects detection using image block percentile color histogram and eigenvector texture feature.
\newblock In {\em First International Conference on Information Sciences, Machinery, Materials and Energy}, pages 779--783. Atlantis Press, 2015.

\bibitem{yuan2015improved}
Xiao-cui Yuan, Lu-shen Wu, and Qingjin Peng.
\newblock An improved otsu method using the weighted object variance for defect detection.
\newblock {\em Applied surface science}, 349:472--484, 2015.

\bibitem{erazo2019histograms}
Jorge Erazo-Aux, H~Loaiza-Correa, and AD~Restrepo-Giron.
\newblock Histograms of oriented gradients for automatic detection of defective regions in thermograms.
\newblock {\em Applied optics}, 58(13):3620--3629, 2019.

\bibitem{cao2024crack}
Jixing Cao, Haijie He, Yao Zhang, Weigang Zhao, Zhiguo Yan, and Hehua Zhu.
\newblock Crack detection in ultrahigh-performance concrete using robust principal component analysis and characteristic evaluation in the frequency domain.
\newblock {\em Structural Health Monitoring}, 23(2):1013--1024, 2024.

\bibitem{wu2018sparse}
Jin-Yi Wu, Stefano Sfarra, and Yuan Yao.
\newblock Sparse principal component thermography for subsurface defect detection in composite products.
\newblock {\em IEEE transactions on industrial informatics}, 14(12):5594--5600, 2018.

\bibitem{liu2020generative}
Kaixin Liu, Yingjie Li, Jianguo Yang, Yi~Liu, and Yuan Yao.
\newblock Generative principal component thermography for enhanced defect detection and analysis.
\newblock {\em IEEE Transactions on Instrumentation and Measurement}, 69(10):8261--8269, 2020.

\bibitem{shakhovska2020improved}
Natalya Shakhovska, Vitaliy Yakovyna, and Natalia Kryvinska.
\newblock An improved software defect prediction algorithm using self-organizing maps combined with hierarchical clustering and data preprocessing.
\newblock In {\em International Conference on Database and Expert Systems Applications}, pages 414--424. Springer, 2020.

\bibitem{tolba1997self}
AS~Tolba and AN~Abu-Rezeq.
\newblock A self-organizing feature map for automated visual inspection of textile products.
\newblock {\em Computers in Industry}, 32(3):319--333, 1997.

\bibitem{mathavan2015use}
Senthan Mathavan, Mujib Rahman, and Khurram Kamal.
\newblock Use of a self-organizing map for crack detection in highly textured pavement images.
\newblock {\em Journal of Infrastructure Systems}, 21(3):04014052, 2015.

\bibitem{cohn2021unsupervised}
Ryan Cohn and Elizabeth Holm.
\newblock Unsupervised machine learning via transfer learning and k-means clustering to classify materials image data.
\newblock {\em Integrating Materials and Manufacturing Innovation}, 10(2):231--244, 2021.

\bibitem{leung2001representing}
Thomas Leung and Jitendra Malik.
\newblock Representing and recognizing the visual appearance of materials using three-dimensional textons.
\newblock {\em International journal of computer vision}, 43:29--44, 2001.

\bibitem{igual2020hierarchical}
Jorge Igual.
\newblock Hierarchical clustering of materials with defects using impact-echo testing.
\newblock {\em IEEE Transactions on Instrumentation and Measurement}, 69(8):5316--5324, 2020.

\bibitem{brzakovic1990approach}
Dragana Brzakovic, H~Beck, and N~Sufi.
\newblock An approach to defect detection in materials characterized by complex textures.
\newblock {\em Pattern Recognition}, 23(1-2):99--107, 1990.

\bibitem{wang2008recognition}
Chih-Hsuan Wang.
\newblock Recognition of semiconductor defect patterns using spatial filtering and spectral clustering.
\newblock {\em Expert Systems with Applications}, 34(3):1914--1923, 2008.

\bibitem{tsai2021autoencoder}
Du-Ming Tsai and Po-Hao Jen.
\newblock Autoencoder-based anomaly detection for surface defect inspection.
\newblock {\em Advanced Engineering Informatics}, 48:101272, 2021.

\bibitem{wang2021variational}
Shuyu Wang, Zhitao Zhong, Yuliang Zhao, and Lei Zuo.
\newblock A variational autoencoder enhanced deep learning model for wafer defect imbalanced classification.
\newblock {\em IEEE Transactions on Components, Packaging and Manufacturing Technology}, 11(12):2055--2060, 2021.

\bibitem{cang2017microstructure}
Ruijin Cang, Yaopengxiao Xu, Shaohua Chen, Yongming Liu, Yang Jiao, and Max Yi~Ren.
\newblock Microstructure representation and reconstruction of heterogeneous materials via deep belief network for computational material design.
\newblock {\em Journal of Mechanical Design}, 139(7):071404, 2017.

\bibitem{liu2019real}
Yumin Liu, Haofei Zhou, Fugee Tsung, and Shuai Zhang.
\newblock Real-time quality monitoring and diagnosis for manufacturing process profiles based on deep belief networks.
\newblock {\em Computers \& Industrial Engineering}, 136:494--503, 2019.

\bibitem{zhang2024maegan}
Y.~Zhang, G.~Yuan, H.~Wu, and H.~Zhou.
\newblock Mae-gan: a self-supervised learning-based classification model for cigarette appearance defects.
\newblock {\em Applied Computing and Information}, 2024.

\bibitem{era2024thermal}
IZ~Era, F~Zhou, AS~Raihan, I~Ahmed, and A~Abul-Haj.
\newblock In-situ melt pool characterization via thermal imaging for defect detection in directed energy deposition using vision transformers.
\newblock {\em arXiv preprint arXiv:2411.12028}, 2024.

\bibitem{yao2024anomaly}
M.~Yao, D.~Tao, P.~Qi, and R.~Gao.
\newblock Rethinking discrepancy analysis: Anomaly detection via meta-learning powered dual-source representation differentiation.
\newblock {\em IEEE Transactions on Automation Science and Engineering}, 2024.

\bibitem{li2024steel}
Z.~Li and J.~Li.
\newblock Single-class detection method for bar material surface defects based on self-supervised learning.
\newblock {\em IEEE Youth Academic Annual Conference Proceedings}, 2024.

\bibitem{hu2024segmentation}
Z.~Hu, H.~Chu, Y.~Zhang, D.~Shan, and Y.~Shen.
\newblock Self-supervised assisted multi-task learning network for one-shot defect segmentation with fake defect generation.
\newblock {\em Pattern Recognition Letters}, 2024.

\bibitem{helwing2024damage}
R.~Helwing, S.~Mrzljak, D.~Hülsbusch, and F.~Walther.
\newblock Cycle-consistent generative adversarial networks for damage evolution analysis in fiber-reinforced polymers based on synthetic damage states.
\newblock {\em Composites Science and Technology}, 2024.

\bibitem{dudzik2015two}
Sebastian Dudzik.
\newblock Two-stage neural algorithm for defect detection and characterization uses an active thermography.
\newblock {\em Infrared Physics \& Technology}, 71:187--197, 2015.

\bibitem{mahmoudi2019layerwise}
Mohamad Mahmoudi, Ahmed~Aziz Ezzat, and Alaa Elwany.
\newblock Layerwise anomaly detection in laser powder-bed fusion metal additive manufacturing.
\newblock {\em Journal of Manufacturing Science and Engineering}, 141(3):031002, 2019.

\bibitem{xing2021convolutional}
Junjie Xing and Minping Jia.
\newblock A convolutional neural network-based method for workpiece surface defect detection.
\newblock {\em Measurement}, 176:109185, 2021.

\bibitem{fang2022tactile}
Bin Fang, Xingming Long, Fuchun Sun, Huaping Liu, Shixin Zhang, and Cheng Fang.
\newblock Tactile-based fabric defect detection using convolutional neural network with attention mechanism.
\newblock {\em IEEE Transactions on Instrumentation and Measurement}, 71:1--9, 2022.

\bibitem{chen2021defect}
Renxiang Chen, Dongyin Cai, Xiaolin Hu, Zan Zhan, and Shuai Wang.
\newblock Defect detection method of aluminum profile surface using deep self-attention mechanism under hybrid noise conditions.
\newblock {\em IEEE Transactions on Instrumentation and Measurement}, 70:1--9, 2021.

\bibitem{liu2024multi}
Jinkang Liu, Xiangyun Long, Chao Jiang, and Wangwang Liao.
\newblock Multi-feature vision transformer for automatic defect detection and quantification in composites using thermography.
\newblock {\em NDT \& E International}, 143:103033, 2024.

\bibitem{wang2023defect}
Junpu Wang, Guili Xu, Fuju Yan, Jinjin Wang, and Zhengsheng Wang.
\newblock Defect transformer: An efficient hybrid transformer architecture for surface defect detection.
\newblock {\em Measurement}, 211:112614, 2023.

\bibitem{pang2024steel}
Wenkai Pang and Zhi Tan.
\newblock A steel surface defect detection model based on graph neural networks.
\newblock {\em Measurement Science and Technology}, 35(4):046201, 2024.

\bibitem{thomas2023materials}
Akhil Thomas, Ali~Riza Durmaz, Mehwish Alam, Peter Gumbsch, Harald Sack, and Chris Eberl.
\newblock Materials fatigue prediction using graph neural networks on microstructure representations.
\newblock {\em Scientific Reports}, 13(1):12562, 2023.

\bibitem{wang2022graph}
Y.~Wang, L.~Gao, Y.~Gao, and X.~Li.
\newblock A graph guided convolutional neural network for surface defect recognition.
\newblock {\em IEEE Transactions on Automation Science and Engineering}, 19(3):1392--1404, July 2022.

\bibitem{hu2019lstm}
Caiqi Hu, Yuxia Duan, Shicai Liu, Yiqian Yan, Ning Tao, Ahmad Osman, Clemente Ibarra-Castanedo, Stefano Sfarra, Dapeng Chen, and Cunlin Zhang.
\newblock Lstm-rnn-based defect classification in honeycomb structures using infrared thermography.
\newblock {\em Infrared Physics \& Technology}, 102:103032, 2019.

\bibitem{wang2020defect}
Qiang Wang, Qiuhan Liu, Ruicong Xia, Guangyuan Li, Jianguo Gao, Hongbin Zhou, and Boyan Zhao.
\newblock Defect depth determination in laser infrared thermography based on lstm-rnn.
\newblock {\em IEEE Access}, 8:153385--153393, 2020.

\bibitem{zhao2024deep}
Oliver Zhao, Dominik Suwito, Bongsub Lee, Thomas Workman, and Laura Mirkarimi.
\newblock Deep convolution neural networks for automatic detection of defects which impact hybrid bonding yield.
\newblock In {\em 2024 IEEE 74th Electronic Components and Technology Conference (ECTC)}, pages 491--497. IEEE, 2024.

\bibitem{zhou2023defect}
Mudan Zhou, Wentao Lu, Jingbo Xia, and Yuhao Wang.
\newblock Defect detection in steel using a hybrid attention network.
\newblock {\em Sensors}, 23(15):6982, 2023.

\bibitem{alipour2020increasing}
Mohamad Alipour and Devin~K Harris.
\newblock Increasing the robustness of material-specific deep learning models for crack detection across different materials.
\newblock {\em Engineering Structures}, 206:110157, 2020.

\bibitem{fujii2016defect}
Hiromitsu Fujii, Atsushi Yamashita, and Hajime Asama.
\newblock Defect detection with estimation of material condition using ensemble learning for hammering test.
\newblock In {\em 2016 IEEE International Conference on Robotics and Automation (ICRA)}, pages 3847--3854. IEEE, 2016.

\bibitem{liu2019data}
Yuekai Liu, Hongli Gao, Liang Guo, Aoping Qin, Canyu Cai, and Zhichao You.
\newblock A data-flow oriented deep ensemble learning method for real-time surface defect inspection.
\newblock {\em IEEE Transactions on Instrumentation and Measurement}, 69(7):4681--4691, 2019.

\bibitem{liu2022defect}
Jiahuan Liu, Fei Guo, Yun Zhang, Binkui Hou, and Huamin Zhou.
\newblock Defect classification on limited labeled samples with multiscale feature fusion and semi-supervised learning.
\newblock {\em Applied Intelligence}, pages 1--16, 2022.

\bibitem{wang2021new}
Yucheng Wang, Liang Gao, Yiping Gao, and Xinyu Li.
\newblock A new graph-based semi-supervised method for surface defect classification.
\newblock {\em Robotics and Computer-Integrated Manufacturing}, 68:102083, 2021.

\bibitem{di2019surface}
He~Di, Xu~Ke, Zhou Peng, and Zhou Dongdong.
\newblock Surface defect classification of steels with a new semi-supervised learning method.
\newblock {\em Optics and Lasers in Engineering}, 117:40--48, 2019.

\bibitem{he2019semi}
Yu~He, Kechen Song, Hongwen Dong, and Yunhui Yan.
\newblock Semi-supervised defect classification of steel surface based on multi-training and generative adversarial network.
\newblock {\em Optics and Lasers in Engineering}, 122:294--302, 2019.

\bibitem{gao2020semi}
Yiping Gao, Liang Gao, Xinyu Li, and Xuguo Yan.
\newblock A semi-supervised convolutional neural network-based method for steel surface defect recognition.
\newblock {\em Robotics and Computer-Integrated Manufacturing}, 61:101825, 2020.

\bibitem{dai2020soldering}
Wenting Dai, Abdul Mujeeb, Marius Erdt, and Alexei Sourin.
\newblock Soldering defect detection in automatic optical inspection.
\newblock {\em Advanced Engineering Informatics}, 43:101004, 2020.

\bibitem{chen2024weakly}
Z.~Chen, L.~Zuo, F.~Guo, C.~Zhang, and Y.~Liu.
\newblock Weakly supervised end-to-end learning for inspection on multidirectional integrated circuit markings in surface mount technology.
\newblock {\em IEEE Transactions on Industrial Informatics}, 20(3):3133--3143, March 2024.

\bibitem{chu2020neural}
Wen-Hsuan Chu and Kris~M Kitani.
\newblock Neural batch sampling with reinforcement learning for semi-supervised anomaly detection.
\newblock In {\em Computer Vision--ECCV 2020: 16th European Conference, Glasgow, UK, August 23--28, 2020, Proceedings, Part XXVI 16}, pages 751--766. Springer, 2020.

\bibitem{chung2022reinforcement}
Jihoon Chung, Bo~Shen, Andrew Chung~Chee Law, and Zhenyu~James Kong.
\newblock Reinforcement learning-based defect mitigation for quality assurance of additive manufacturing.
\newblock {\em Journal of Manufacturing Systems}, 65:822--835, 2022.

\bibitem{zhang2023duak}
Yufei Zhang, Hongwei Wang, Weiming Shen, and Gongzhuang Peng.
\newblock Duak: Reinforcement learning-based knowledge graph reasoning for steel surface defect detection.
\newblock {\em IEEE Transactions on Automation Science and Engineering}, 2023.

\bibitem{mohanty2024enhancing}
Sankarsan Mohanty, Eugene Su, and Chao-Ching Ho.
\newblock Enhancing titanium spacer defect detection through reinforcement learning-optimized digital twin and synthetic data generation.
\newblock {\em Journal of Electronic Imaging}, 33(1):013021--013021, 2024.

\bibitem{zhou2023comparative}
Qianqian Zhou, Zuxiang Situ, Shuai Teng, and Gongfa Chen.
\newblock Comparative effectiveness of data augmentation using traditional approaches versus stylegans in automated sewer defect detection.
\newblock {\em Journal of Water Resources Planning and Management}, 149(9):04023045, 2023.

\bibitem{huang2020surface}
Yibin Huang, Congying Qiu, and Kui Yuan.
\newblock Surface defect saliency of magnetic tile.
\newblock {\em The Visual Computer}, 36:85--96, 2020.

\bibitem{dabetwar2022fatigue}
Shweta Dabetwar, Stephen Ekwaro-Osire, and Jo{\~a}o~Paulo Dias.
\newblock Fatigue damage diagnostics of composites using data fusion and data augmentation with deep neural networks.
\newblock {\em Journal of Nondestructive Evaluation, Diagnostics and Prognostics of Engineering Systems}, 5(2):021004, 2022.

\bibitem{valente2020print}
Augusto Valente, Cristina Wada, Deangela Neves, Deangeli Neves, Fabio Perez, Guilherme Megeto, Marcos Cascone, Otavio Gomes, and Qian Lin.
\newblock Print defect mapping with semantic segmentation.
\newblock In {\em Proceedings of the IEEE/CVF winter conference on applications of computer vision}, pages 3551--3559, 2020.

\bibitem{bosse2023automated}
Stefan Bosse.
\newblock Automated damage and defect detection with low-cost x-ray radiography using data-driven predictor models and data augmentation by x-ray simulation.
\newblock {\em Eng. Proc}, 56, 2023.

\bibitem{farady2023preaugnet}
Isack Farady, Chih-Yang Lin, and Ming-Ching Chang.
\newblock Preaugnet: improve data augmentation for industrial defect classification with small-scale training data.
\newblock {\em Journal of Intelligent Manufacturing}, pages 1--14, 2023.

\bibitem{yun2020automated}
Jong~Pil Yun, Woosang~Crino Shin, Gyogwon Koo, Min~Su Kim, Chungki Lee, and Sang~Jun Lee.
\newblock Automated defect inspection system for metal surfaces based on deep learning and data augmentation.
\newblock {\em Journal of Manufacturing Systems}, 55:317--324, 2020.

\bibitem{pourreza2021g2d}
Masoud Pourreza, Bahram Mohammadi, Mostafa Khaki, Samir Bouindour, Hichem Snoussi, and Mohammad Sabokrou.
\newblock G2d: Generate to detect anomaly.
\newblock In {\em Proceedings of the IEEE/CVF Winter Conference on Applications of Computer Vision}, pages 2003--2012, 2021.

\bibitem{zhang2021defect}
Gongjie Zhang, Kaiwen Cui, Tzu-Yi Hung, and Shijian Lu.
\newblock Defect-gan: High-fidelity defect synthesis for automated defect inspection.
\newblock In {\em Proceedings of the IEEE/CVF Winter Conference on Applications of Computer Vision}, pages 2524--2534, 2021.

\bibitem{jain2022synthetic}
Saksham Jain, Gautam Seth, Arpit Paruthi, Umang Soni, and Girish Kumar.
\newblock Synthetic data augmentation for surface defect detection and classification using deep learning.
\newblock {\em Journal of Intelligent Manufacturing}, pages 1--14, 2022.

\bibitem{wang2024}
Jianjun Wang and Mei Liu.
\newblock Enhanced yolov8 for precise steel defect detection on neu dataset.
\newblock {\em Materials Today Communications}, 50:101780, 2024.

\bibitem{zhu2024mrp}
Shuxian Zhu and Yajie Zhou.
\newblock Mrp-yolo: An improved yolov8 algorithm for steel surface defects.
\newblock {\em Machines}, 12(12):917, 2024.

\bibitem{zhu2015autocorrelation}
Yong Zhu and Lili Li.
\newblock Fabric defect detection using autocorrelation and glcm features.
\newblock {\em Textile Research Journal}, 85(4):435--445, 2015.

\bibitem{shalaby2024condition}
Yasmin~M Shalaby, Mohamed Badawy, Gamal~A Ebrahim, and Ahmed~Mohammed Abdelalim.
\newblock Condition assessment of concrete structures using automated crack detection method for different concrete surface types based on image processing.
\newblock {\em Discover Civil Engineering}, 1(1):81, 2024.

\bibitem{li2021anomaly}
Ning Li, Kaitao Jiang, Zhiheng Ma, Xing Wei, Xiaopeng Hong, and Yihong Gong.
\newblock Anomaly detection via self-organizing map.
\newblock In {\em 2021 IEEE International Conference on Image Processing (ICIP)}, pages 974--978. IEEE, 2021.

\bibitem{tao2023vitalnet}
Xian Tao, Chandranath Adak, Pang-Jo Chun, Shaohua Yan, and Huaping Liu.
\newblock Vitalnet: Anomaly on industrial textured surfaces with hybrid transformer.
\newblock {\em IEEE Transactions on Instrumentation and Measurement}, 72:1--13, 2023.

\bibitem{olimov2022uzadl}
BA~Ugli Olimov, KC~Veluvolu, and A~Paul.
\newblock Uzadl: Anomaly detection and localization using graph laplacian matrix-based unsupervised learning method.
\newblock {\em Computers \& Industrial Engineering}, 165:107950, 2022.

\bibitem{elsken2019neural}
Thomas Elsken, Jan~Hendrik Metzen, and Frank Hutter.
\newblock Neural architecture search: A survey.
\newblock {\em Journal of Machine Learning Research}, 20(1):1--21, 2019.

\bibitem{liu2021survey}
Hanxiao Liu, Karen Simonyan, and Yiming Yang.
\newblock A comprehensive survey on neural architecture search.
\newblock {\em IEEE Transactions on Pattern Analysis and Machine Intelligence}, 44(9):7289--7319, 2021.

\bibitem{hinton2015distilling}
Geoffrey Hinton, Oriol Vinyals, and Jeff Dean.
\newblock Distilling the knowledge in a neural network.
\newblock {\em Advances in Neural Information Processing Systems}, 28, 2015.

\bibitem{xu2020edge}
Xiaofei Xu, Jie Zhang, and Wenjing Yang.
\newblock Edge computing: A survey on the state of the art.
\newblock {\em Frontiers of Computer Science}, 14:1--23, 2020.

\bibitem{kang2017neurosurgeon}
Youngjae Kang, Johann Hauswald, Chunjing Gao, Andrew Rovinski, Trevor Mudge, Jason Mars, and Lingjia Tang.
\newblock Neurosurgeon: Collaborative intelligence between the cloud and mobile edge.
\newblock {\em ACM SIGARCH Computer Architecture News}, 45(1):615--629, 2017.

\bibitem{osman2020automatic}
S.~A. Osman, J.~Abdullah, and M.~Mustapha.
\newblock Automatic defect detection using texture analysis methods: A review.
\newblock {\em Journal of Advanced Research}, 25:102--117, 2020.

\bibitem{hui2019application}
W.~Hui, Z.~Xu, and M.~Lin.
\newblock Application of glcm in composite material defect detection.
\newblock {\em Journal of Materials Processing Technology}, 264:122--130, 2019.

\bibitem{li2018gabor}
X.~Li, G.~Sun, and Y.~Zhang.
\newblock Gabor filter-based texture feature extraction for composite defect detection.
\newblock {\em Composite Structures}, 182:162--171, 2018.

\bibitem{sharma2021grey}
S.~Sharma, R.~Kumar, and A.~Jain.
\newblock Grey-level co-occurrence matrix-based texture analysis for defect identification in composites.
\newblock {\em Materials Today: Proceedings}, 45:3072--3081, 2021.

\bibitem{vasan2019defect}
A.~S. Vasan, B.~Surendiran, and M.~Balaji.
\newblock Defect detection in composite materials using glcm features.
\newblock {\em Journal of Non-Destructive Evaluation}, 38(2):1--10, 2019.

\bibitem{mcknight20243}
Shaun McKnight, Christopher MacKinnon, S~Gareth Pierce, Ehsan Mohseni, Vedran Tunukovic, Charles~N MacLeod, Randika~KW Vithanage, and Tom O’Hare.
\newblock 3-dimensional residual neural architecture search for ultrasonic defect detection.
\newblock {\em IEEE Transactions on Ultrasonics, Ferroelectrics, and Frequency Control}, 2024.

\bibitem{oztacs2024image}
Burak {\"O}zta{\c{s}}, Yasemin Korkmaz, and Halil~{\.I}brahim {\c{C}}elik.
\newblock Image analyses of artificially damaged carbon/glass/epoxy composites before and after impact load.
\newblock {\em Heliyon}, 2024.

\bibitem{lu2024research}
Zhaoxu Lu, Kai Yao, Xinglong Li, and Chenghao Yu.
\newblock Research on ultrasonic defect imaging based on a neural network with gaussian weight function fusion model.
\newblock {\em Construction and Building Materials}, 411:134229, 2024.

\bibitem{gong2020deep}
Yanfeng Gong, Hongliang Shao, Jun Luo, and Zhixue Li.
\newblock A deep transfer learning model for inclusion defect detection of aeronautics composite materials.
\newblock {\em Composite structures}, 252:112681, 2020.

\bibitem{zhang2024one}
Yubin Zhang, Changhang Xu, Pengqian Liu, Jing Xie, Yage Han, Rui Liu, and Lina Chen.
\newblock One-dimensional deep convolutional autoencoder active infrared thermography: Enhanced visualization of internal defects in frp composites.
\newblock {\em Composites Part B: Engineering}, 272:111216, 2024.

\bibitem{deng2024attention}
Kailun Deng, Haochen Liu, Jun Cao, Lichao Yang, Weixiang Du, Yigeng Xu, and Yifan Zhao.
\newblock Attention mechanism enhanced spatiotemporal-based deep learning approach for classifying barely visible impact damages in cfrp materials.
\newblock {\em Composite Structures}, 337:118030, 2024.

\bibitem{palka2024automatic}
Norbert Pa{\l}ka, Kamil Kami{\'n}ski, Marcin Maciejewski, Krzysztof Dragan, and Piotr Synaszko.
\newblock Automatic histogram-based defect detection in glass fibre reinforced polymer composites using terahertz time-domain spectroscopy reflection imaging.
\newblock {\em Optics and Lasers in Engineering}, 174:107959, 2024.

\bibitem{mahmoud2024enhancing}
Alaaeldin Mahmoud, Mohammed Kassem, Ahmed Elrewainy, and Yasser~H El-Sharkawy.
\newblock Enhancing automatic inspection and characterization of carbon fiber composites through hyperspectral diffuse reflection analysis and k-means clustering.
\newblock {\em The International Journal of Advanced Manufacturing Technology}, pages 1--17, 2024.

\bibitem{Vavilov2020}
V.~P. Vavilov, A.~A. Karabutov, A.~O. Chulkov, D.~A. Derusova, A.~I. Moskovchenko, E.~B. Cherepetskaya, and E.~A. Mironova.
\newblock Comparative study of active infrared thermography, ultrasonic laser vibrometry and laser ultrasonics in application to the inspection of graphite/epoxy composite parts.
\newblock {\em Quantitative InfraRed Thermography Journal}, 17(4):235--248, 2020.

\bibitem{sudharsan2024multi}
PL~Sudharsan, Thulsiram Gantala, and Krishnan Balasubramaniam.
\newblock Multi modal data fusion of paut with thermography assisted by automatic defect recognition system (m-adr) for nde applications.
\newblock {\em NDT \& E International}, 143:103062, 2024.

\bibitem{ye2023synchronous}
Chaofeng Ye, Haoran Dong, and Na~Zhang.
\newblock Synchronous imaging and multimodal fusion of optical and electromagnetic measurements for overlapping defects inspection.
\newblock {\em IEEE Transactions on Industrial Informatics}, 19(2):1543--1554, 2023.

\end{thebibliography}

\end{document}